# WhONet: Wheel Odometry Neural Network for Vehicular Localisation in GNSS-Deprived Environments

Uche Onyekpe[1,4*], Vasile Palade[1], Anuradha Herath[2], Stratis Kanarachos[3] and Michael E. Fitzpatrick[2,3]

[1] *Research Centre for Data Science, Coventry University, Priory Road, Coventry, CV1 5FB, United Kingdom*

[2] *Institute for Future Transport and Cities, Coventry University, Gulson Road, Coventry, CV1 5FB, United Kingdom*

[3] *Faculty of Engineering, Coventry University, Priory Road, Coventry, CV1 5 FB, United Kingdom*

[4] *Department of Data Science, York St John University, York YO31 7EX, United Kingdom*

Email: u.onyekpe@yorksj.ac.uk, ab5839@coventry.ac.uk, herathh@uni.coventry.ac.uk, ab8522@coventry.ac.uk, ab6856@coventry.ac.uk

**Abstract**

In this paper, a deep learning approach is proposed to accurately position wheeled vehicles in Global Navigation Satellite Systems (GNSS) deprived environments. In the absence of GNSS signals, information on the speed of the wheels of a vehicle (or other robots alike), recorded from the wheel encoder, can be used to provide continuous positioning information for the vehicle, through the integration of the vehicle's linear velocity to displacement. However, the displacement estimation from the wheel speed measurements are characterised by uncertainties, which could be manifested as wheel slips or/and changes to the tyre size or pressure, from wet and muddy road drives or tyres wearing out. As such, we exploit recent advances in deep learning to propose the **Wh**eel **O**dometry neural **Net**work (WhONet) to learn the uncertainties in the wheel speed measurements needed for correction and accurate positioning. The performance of the proposed WhONet is first evaluated on several challenging driving scenarios, such as on roundabouts, sharp cornering, hard-brake and wet roads (drifts). WhONet's performance is then further and extensively evaluated on longer-term GNSS outage scenarios of 30s, 60s, 120s and 180s duration, respectively over a total distance of 493 km. The experimental results obtained show that the proposed method is able to accurately position the vehicle with up to 93% reduction in the positioning error of its original counterpart (physics model) after any 180s of travel.

Keywords: Wheel odometry, Autonomous vehicles, Inertial Navigation System, Deep learning, Machine learning, GNSS outage, Positioning, Neural networks

## 1. Introduction

The positioning and orientation of autonomous vehicles and robots alike are needed for decision making and navigation purposes. Several approaches based on different sensor combinations have been proposed to track the motion of the vehicles in situations where satellite signals are not available for localisation: so-called GNSS-deprived environments. Those sensors include low-cost Inertial Navigation System (INS) sensors, which commonly consists of an accelerometer to measure the acceleration of the vehicle for displacement estimation, and a gyroscope to measure the vehicles attitude rate for orientation estimation. The measurements of the INS are however affected by noises (such as thermo-mechanical noise, flicker noise, calibration errors, etc.), which are amplified during the double integration of the vehicles acceleration to displacement and orientation rate to orientation and are cascaded exponentially over time. This error accumulation is particularly due to the incremental position estimation process of the Dead Reckoning (DR) approach to INS positioning. Therefore, continuous updating to correct the accumulated error is required.

Several techniques based on machine learning have been investigated to learn the error drift over time to provide an accurate estimation of the vehicles position and orientation. According to the relevant literature, there is a huge potential for machine learning based algorithms in handling navigation in autonomous systems. Malleswaran et.al. in [1], studied the use of multiplicative neural networks and sigma pi neural network on vehicular positioning using INS/GPS (Global Positioning System) integrated systems. The Input Delay Neural Network (IDNN) [2], was explored by Noureldin et.al. on the ability to learn the positioning error in GPS outages. Chiang in [3] explored the use of a Multi-layer Feed Forward Neural Network (MFNN) in positioning using a single point INS/GPS architecture. A position estimation solution was proposed by Sharaf et al. in [4] using MFNN on an integrated INS (tactical grade) and a differential GPS architecture. In [5], Malleswaran et al. comparatively investigated the use of the forward-only counter propagation neural network, Radial Basis Function Neural Network (RBFNN), Full Counter Propagation Neural network (Full CPN), Higher-Order Neural Networks (HONN), back propagation neural network, Adaptive Resonance Theory-Counter Propagation Neural network (ART-CPN), and the IDNN to predict the position error of the INS during Global Positioning System (GPS) outages. Noureldin et al. in [6] investigated the use of a neuro fuzzy augmented Kalman filter to improve the stochastic modelling of the Micro-Electro-Mechanical Systems (MEMS)-based inertial errors. Malleswaran et al. in [7] studied the use of recurrent neural networks with evolutionary optimization techniques for the integration of INS and GPS. In [8], El-Sheimy et al. proposed an MFNN with a conjugate gradient training algorithm to fuse uncompensated INS measurements and differential GPS measurement. A radial basis function neural network (RBFNN) was used by Semeniuk and Noureldin in [9] to predict the position and velocity errors of a tactical grade INS integrated with GPS and showed the superiority of the RBFNN over the Kalman Filter. Chiang et al. in [10] and Malleswaran et al. in [11] demonstrated the improved performance of constructive neural networks over MFNNs for INS/GPS integration. Dai et al. [12] investigated the use of the classic Recurrent Neural Networks (RNN) in modelling the time dependant error drift of INS. RNNs are different from other neural networks as they have the ability to learn relationships in temporal sequences. A performance comparison of LSTM algorithm and Multi-layer Neural Network was done by Fang et al. in [13]. Their research clearly suggests that Long Short-Term Memory (LSTM) algorithm is better in performance than Multi-Layer Perceptron (MLP) on ground vehicular positioning. Onyekpe et al. in [14] explored the performance of the LSTM in comparison to other techniques like the IDNN, MLP, and Kalman filter for high data rate positioning. In [15], they further analysed the performance of the IDNN, LSTM, Gated Recurrent Unit (GRU), RNN and MLP in challenging driving scenarios consisting of hard braking, roundabout, sharp cornering and successive left and right turns. Despite the numerous investigations into improving the performance of low-cost INS, the problem still remains a challenge in need of inexpensive alternate solutions.

* Corresponding author Uche Onyekpe: u.onyekpe@yorksj.ac.uk

Embedded in modern vehicles are a good number of sensors supporting advanced driverless systems, such as the wheel encoder for anti-lock braking systems. The wheel encoder operates by measuring the speed of the vehicle's axle or wheel. However, wheel odometry can also accumulate errors due to the internal systematic factors and non-systematic factors [16]–[18]. Borenstein et al. in [17] based their experiments on systematic errors using differential drive robots and came to the conclusion that "unequal wheel diameters", "uncertainty about the wheelbase", and the "difference between the actual and the nominal wheel diameters" are the most common and important systematic error sources. But according to Borenstein et al. [17], these parameters do not cause errors in all driving scenarios. For instance, unequal wheel diameter causes error in straight line motion and wheelbase uncertainty causes error only when turning. Borenstein et.al. further reveal that non-systemics errors, such as travel on uneven road surfaces and wheel slippage due to wet and slippery roads, over-acceleration, fast turning (skidding), etc. can also cause uncertainties in wheel odometry. While addressing wheel odometry errors from differences in actual and nominal tyre sizes, Onyekpe et al. in [19] revealed that the use of a different tyre pressure than the recommended pressure or replacing tyres with a slightly different tyre diameter can lead to errors in vehicle displacements in odometry. A larger tyre diameter will overestimate the displacement while a smaller tyre diameter can underestimate the vehicle displacement.

However, wheel odometry is better for positioning estimation compared to the accelerometer of the INS. It requires fewer number of integration steps to determine the position of a vehicle, thus minimising the error propagation and making it a desirable and simpler approach. In [19], Onyekpe et al. showed that the Long Short-Term Memory (LSTM) neural network can be used to learn the errors or uncertainties present within the displacement estimation from the wheel speed measurement to provide a more accurate solution compared to that of the INS's accelerometer. We proposed for the first time in [19], an end-to-end neural network-based technique for estimating position errors for vehicular localisation in GNSS-deprived environments. The work done in [19] is further extended here by proposing a novel neural network framework called Wheel Odometry Neural Network (WhONet) for a more accurate positioning of wheeled autonomous vehicles in GNSS-deprived environments. We further extend the performance analysis from the challenging scenarios evaluated in [19], to cover longer GNSS outages consisting of 129 sequences of 180s seconds length each, 197 sequences of 120s seconds length each, 399 sequences of 60s seconds length each, 809 sequences of 30s seconds length each. The evaluation is done using the Inertial and Odometry Vehicle Navigation Benchmark Dataset (IO-VNBD) [20] and the Coventry University Public road dataset for Automated Cars (CUPAC) [21] with analysis over different journey domains. The main contributions and novelties of this paper are thus summarised as highlighted below.

1. A novel neural network approach is proposed for an end-to-end prediction of the positional errors of autonomous vehicles using wheel encoders in GNSS deprived environments.
2. The proposed method is a more robust, accurate and efficient approach compared to [19] in vehicular localisation using wheel encoders.
3. A robust performance analysis on challenging driving scenarios in both short term and long term GNSS outages is done for the first time to the best of our knowledge whilst using wheel odometry data.

The rest of the paper is structured as follows; Section 2 describes the vehicular navigation problems used in evaluating the performance of WhONet as well as the mathematical formulation of the learning problem and the learning scheme of WhONet. In section 3, the datasets and metrics used in evaluating the performance WhONet are defined. A comparative analysis justifying the selection of RNN for use in WhONet as well as the training and test information are also provided in section 3. In section 4, the results obtained from the evaluation of WhONet are discussed and the paper is finally concluded in section 5.

## 2. Problem Description and Mathematical Formulation of the Learning Problem

### 2.1 Vehicular Localisation in Challenging Driving Environments Problem

The problem of vehicular localisation in challenging driving environments was first addressed in [15]. In reference [15], [19] and [22], the performance of several positioning techniques was evaluated on various dynamic conditions such as sharp cornering, roundabouts, and hard brake situations, and proved to be quite challenging for the techniques investigated. However, as many studies on vehicular localisation focus more on non-complex driving conditions which are not consistent with everyday driving, it is important that the performance of the proposed technique is evaluated on such complex scenarios as discussed below.

Hard brake – Hard brake occurs when a larger brake force is applied causing a rapid deceleration. This rapid deceleration makes it difficult to accurately predict the vehicle's motion [15]. Reference [23] defines hard brakes as longitudinal decelerations $\leq -0.45$ g.

Wet road – When a vehicle is driven over a wet/muddy road, the wheels tend to slip whilst driving. This introduces uncertainties to the wheel speed estimation, leading to errors in the displacement calculation of the vehicle.

Changes in acceleration – Changes to the acceleration of the vehicle within a short period of time can be challenging to the accurate tracking of the vehicle's position, as positioning algorithms struggle to accurately capture the rapid changes in displacements and accelerations [10].

Sharp Cornering and Successive Left-Right Turns – Sharp and continuous consecutive changes to the vehicles direction of travel poses a challenge for an accurate tracking of the vehicles motion using INS.

Roundabout – The motion of vehicles around roundabouts presents a struggle particularly due to the roundabout's shape. The displacement and the orientation of the vehicle changes with time as the vehicle turns around the roundabout. Due to the continuous change in the vehicle's direction, it becomes a challenge to accurately predict the positioning of the vehicle. According to investigations in [15], the roundabout is the most challenging of all scenarios discussed.

### 2.2 Longer-term GNSS outage

The problem of vehicular localisation in longer-term GNSS outages (defined as outages up to 180 seconds) is particularly important for navigation in tunnels and valleys, but also finds relevance in other GNSS outage scenarios such as navigation under bridges, through dense tree canopies, urban canyons, etc. The longer-term GNSS outage scenarios studied are characterised by a mixture of complex scenarios as described above, and more as could be found in everyday driving. To assess the relevance of WhONet for vehicular positioning in everyday navigation in longer-term GNSS absence, its performance is evaluated over a total distance of 477 km in assumed GNSS outages of 30s, 60s, 120s and 180s.

2.3 Physics Model: Position Tracking Through the Dead Reckoning of the Wheel Speed Measurement from Newtonian mechanics.

A reference is needed when positional tracking of vehicles is performed using dead reckoning. The measurements are usually provided in the body (sensors) frame and are then transformed into the navigation frame for positioning using the transformation matrix presented in Equation (1) [24].

$$R^{nb} = \begin{bmatrix} cos\Psi & -sin\Psi & 0 \\ sin\Psi & cos\Psi & 0 \\ 0 & 0 & 1 \end{bmatrix} \tag{1}$$

Where $\Psi$ represents the yaw of the vehicle. As this study focuses on navigation on the horizontal plane, the pitch and the roll of the vehicle are neglected.

The angular velocity of the wheels at a given time (t) is measured by the wheel speed sensors. But there can be uncertainties in the tyre's diameter due to the condition of the tyre (such as tyre wearing), tyre pressure, and wheel slip. These uncertainties affect the accuracy of displacement estimation from the wheel speed measurement $\omega$. Equations (2) - (5) consider the errors which can affect the vehicle speed calculations.

$$\widehat{\omega}^b_{whrl} = \omega^b_{whrl} + \varepsilon^b_{whrl} \tag{2}$$

$$\widehat{\omega}^b_{whrr} = \omega^b_{whrr} + \varepsilon^b_{whrr} \tag{3}$$

$$\widehat{\omega}^b_{whfl} = \omega^b_{whfl} + \varepsilon^b_{whfl} \tag{4}$$

$$\widehat{\omega}^b_{whfr} = \omega^b_{whfr} + \varepsilon^b_{whfr} \tag{5}$$

Where $\widehat{\omega}^b_{whrl}$, $\widehat{\omega}^b_{whrr}$, $\widehat{\omega}^b_{whfl}$ and $\widehat{\omega}^b_{whfr}$ are the noisy wheel speed measurements of the rear left, rear right, front left and front right wheels, whereas $\varepsilon^b_{whrl}$, $\varepsilon^b_{whrr}$, $\varepsilon^b_{whfl}$ and $\varepsilon^b_{whfr}$ are the corresponding errors (uncertainties), and $\omega^b_{whrl}$, $\omega^b_{whrr}$, $\omega^b_{whfl}$ and $\omega^b_{whfr}$ are the respective error-free wheel speed measurements.

Equations (6) and (7) show the calculation of the angular velocity (wheel speed) of the rear axle.

$$\widehat{\omega}^b_{whr} = \frac{\omega^b_{whrr} + \omega^b_{whrl}}{2} + \frac{\varepsilon^b_{whrr} + \varepsilon^b_{whrl}}{2} \tag{6}$$

Expressing $\frac{\varepsilon^b_{whrr}+\varepsilon^b_{whrl}}{2}$ as $\varepsilon^b_{whr}$ and $\frac{\omega^b_{whrr}+\omega^b_{whrl}}{2}$ as $\omega^b_{whr}$

$$\widehat{\omega}^b_{whr} = \omega^b_{whr} + \varepsilon^b_{whr} \tag{7}$$

From $v = \omega r$, the linear velocity of the vehicle in the body frame can be found, with $r$ as a constant which maps the wheel speed of the rear axle to the linear velocity of the vehicle:

$$v^b_{wh} = \omega^b_{whr} r + \varepsilon^b_{whr} r \tag{8}$$

Take $\varepsilon^b_{whr} r \ as \ \varepsilon^b_{whr,v}$

$$v^b_{whr} = \omega^b_{whr} r + \varepsilon^b_{whr,v} \tag{9}$$

The displacement of the vehicle in the body frame can be found through the integration of the vehicle's velocity from $Equation$ 9 and incrementally updated for continuous tracking. $\varepsilon^b_{whr,x}$ in $Equation$ 10 is the integral of $\varepsilon^b_{whr,v}$ from $Equation$ 9.

$$x^b_{whr} = \int_{t-1}^{t} (\omega^b_{whr} r) + \varepsilon^b_{whr,x} \tag{10}$$

The uncertainty in the position estimation can be found through $Equation$ (11) during the presence of the GNSS signal. The task thus becomes that of estimating $\varepsilon^b_{whr,x}$ during GNSS outages needed to correct the vehicles displacement $x^b_{whr}$.

$$\varepsilon^b_{whr,x} \approx x^b_{whr} - x^b_{GNSS} \tag{11}$$

$x^b_{GNSS}$ is the vehicle's true displacement measured according to [14] using Vincenty's formula for geodesics on an ellipsoid based on the longitudinal and latitudinal positional information of the vehicle as in [25], [26]. The accuracy of $x^b_{GNSS}$ is however limited to the accuracy of the GNSS which is defined as $\pm$3m according to [27].

## 2.4 WhONet's Learning Scheme

The proposed prediction block of the WhONet approach[1] for the positioning of vehicles in the absence of GNSS signals using the wheel speed information is presented in Figure 1, where NED is the North-East-Down coordinate defining the navigation frame. WhONet uses a classic Recurrent Neural Network (RNN) as justified in section 3.3.

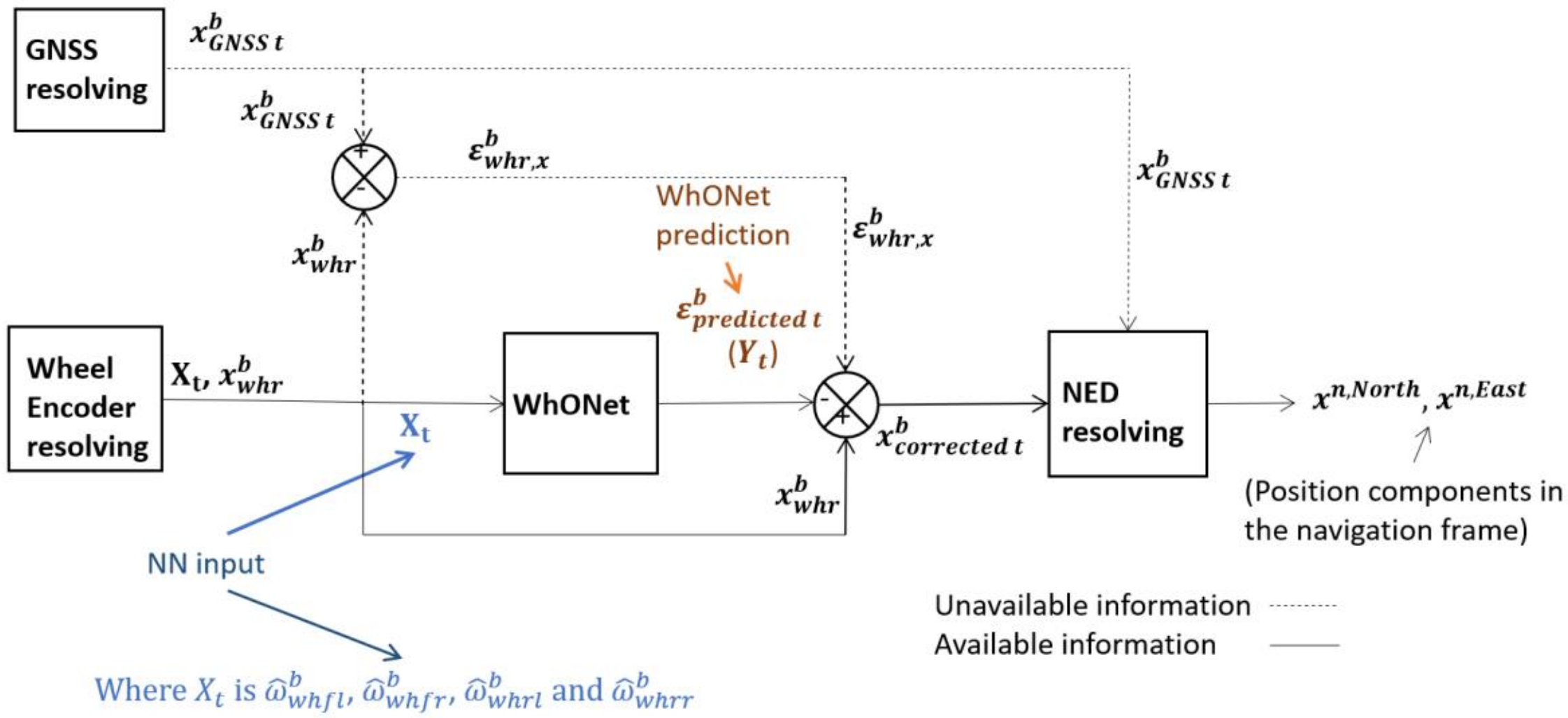

***Figure 1**. The position prediction block during GNSS outages using the WhONet*

The RNN proposed by Rumelhart et al. in [28] has found use in several time series applications. The RNN is uniquely characterised by its unique ability to learn relationships within sequential data. It is made up of feedback loops which connect relationships learnt in the past. These connections are sometimes called memory. The information learnt within the sequential dimension of the data are stored within the hidden state of the RNN and extended across the defined number of time steps before being mapped to the output. Figure 2 shows the unrolled architecture of the classic RNN.

The operations of the RNN are governed by the equations below.

$$h_t = tanh(U_h h_{t-1} + W_x X_t + b_h) \quad (12)$$

$$y_t = \sigma(W_o h_t + b_o) \quad (13)$$

We define $X_t$ as the input feature vector to the RNN, y as the output vector, $\boldsymbol{\sigma}$ as the sigmoid activation (non-linearity) function, $h_{t-1}$ as the previous state, $U_h$ as the hidden weight matrix, $W_x$ as the input weight matrix, $W_o$ as the output weight matrix and $b$ as the bias.

[1] WhONet's implementation can be found at https://github.com/onyekpeu/WhONet.

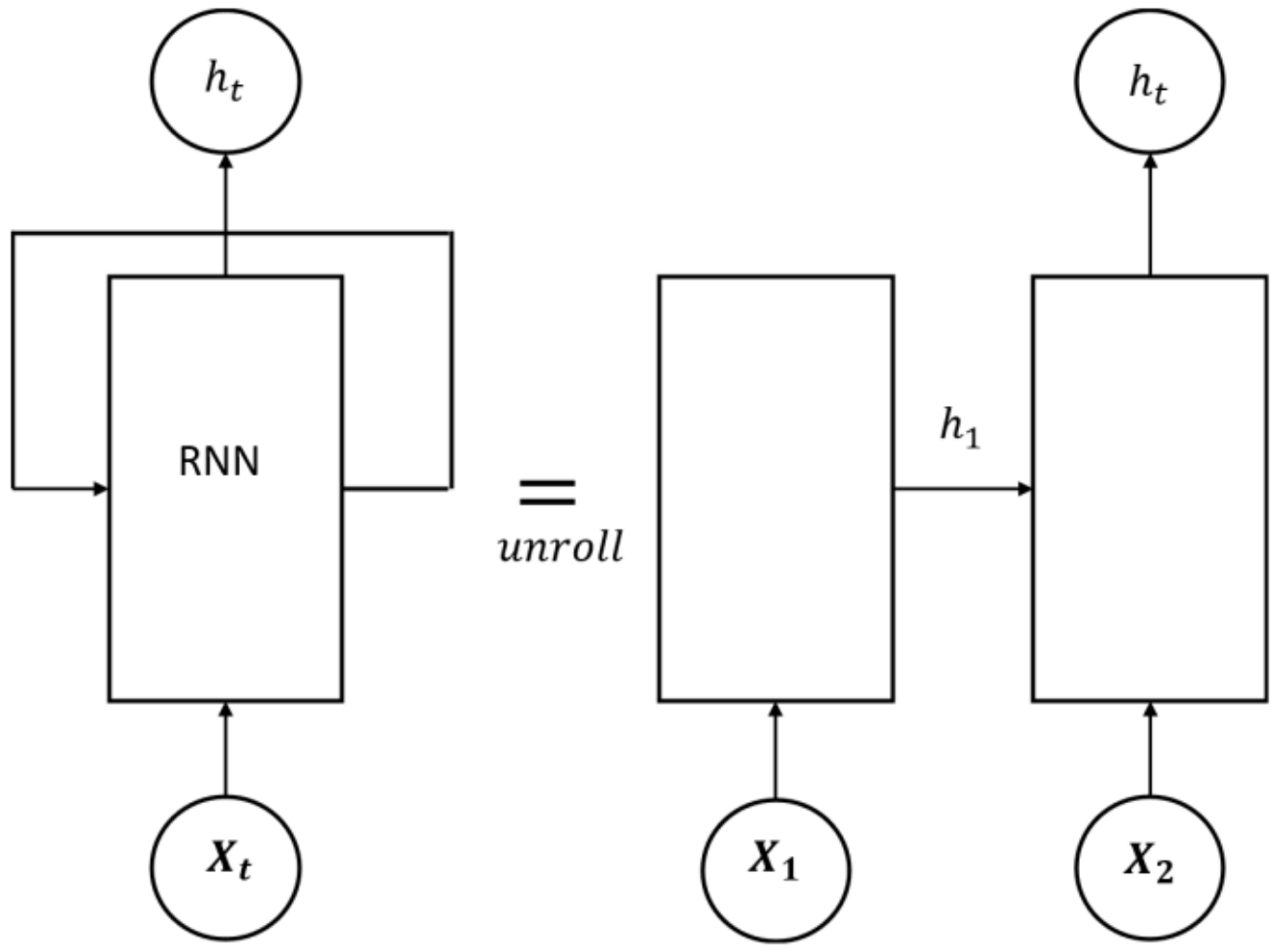

***Figure 2.*** *Unrolled architecture of the RNN used in WhONet*

At any time, $t$:

$$X_1 = X_{t|t-0.9} \tag{14}$$

$$X_2 = X_{t-1|t-1.9} \tag{15}$$

Here $X_t$ is further described as the temporal-sequenced input feature to the RNN at time t, which is characterised by the wheel speed information from the four wheels of the vehicles: $\widehat{\omega}^b_{whrl}$, $\widehat{\omega}^b_{whrr}$, $\widehat{\omega}^b_{whfl}$ and $\widehat{\omega}^b_{whfr}$ and, $X_1$ and $X_2$ are the features at each timestep respectively as shown in Figure 2. The RNN is tasked with predicting $Y_t$, which is the error $\varepsilon^b_{whr,x}$ between the wheel-speed-derived displacement $x^b_{whr}$ and the GNSS-derived displacement $x^b_{GNSS}$.

The differences between the WhONet model and the approach presented in [19] are presented in Table 1.

Table1. Comparison between the WhONet model and the method presented in [19].

| No | WhONet (proposed approach) | [19] |
|---|---|---|
| 1 | The RNN architecture is structured to accommodate more information on the vehicles motion state i.e., higher sampled wheel odometry data at 10Hz | The LSTM architecture is structured to accommodate low sampled wheel odometry data at 1Hz |
| 2 | More accurate position estimation compared to [19]. | Less accurate position estimations compared to WhONet. |
| 3 | Relatively more computationally efficient compared to [19] due to fewer parameterisation of the RNN and no gated operations. | Relatively less computationally efficient compared to WhONet due to more parameterisation of the LSTM as well as its gated operations. |

## 3. Experiment and Dataset

### 3.1 Dataset

This study uses two datasets, the IO-VNBD and the CUPAC, as discussed below.

#### 3.1.1 IO-VNBD (Inertial and Odometry Vehicle Navigation Benchmark Dataset)

The IO-VNBD dataset is characterised by a diverse number of driving scenarios such as hard brakes, roundabout, town drives, residential road drives, dirt roads, traffic, sharp cornering, etc., and was collected over 5700km and 98 hours of driving. A Ford Fiesta Titanium (as shown in Figure 3) was used as the vehicle for the data collection on public roads in the United Kingdom. The dataset consists of information of the vehicle's dynamics and position such as the vehicle's wheel speeds (in rad/sec) and GPS coordinates (in degrees) extracted from the Electronic Control Unit (ECU) of the vehicle at a sampling rate of 10 Hz. More details of the dataset can be found in [20].

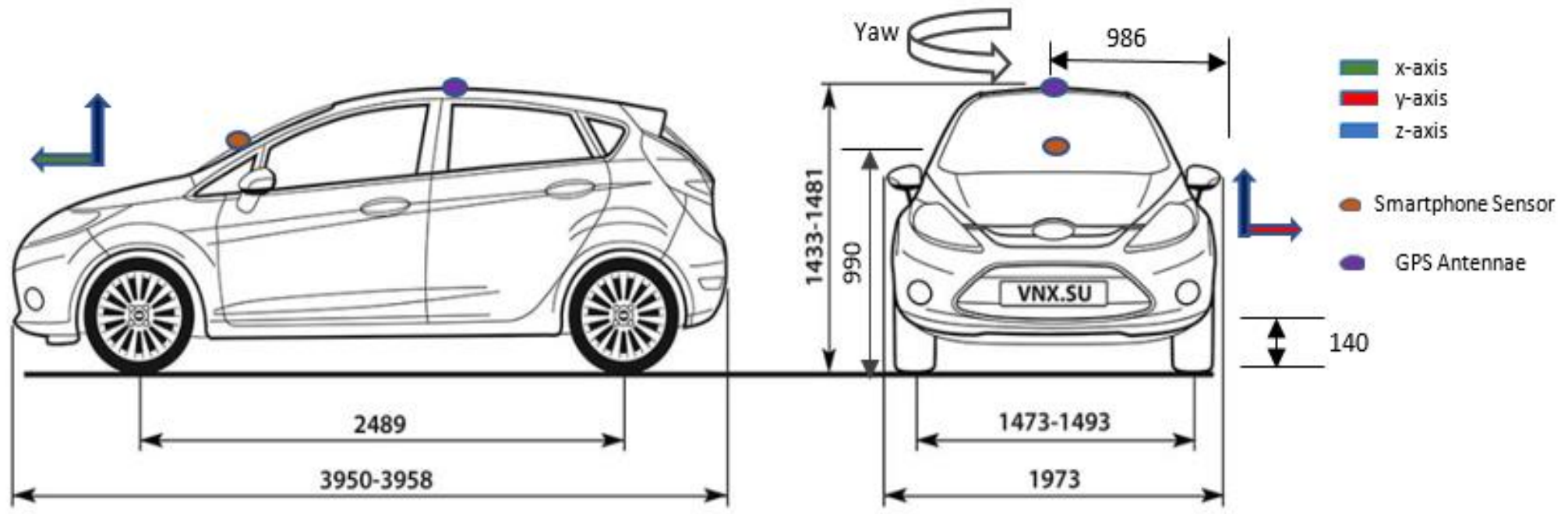

***Figure 1.*** *Data collection vehicle, showing sensor locations* [20].

3.1.2 CUPAC (Coventry University Public road dataset for Automated Cars)

This CUPAC dataset was collected on public streets in Coventry, United Kingdom under different weather, traffic and road conditions. A LIDAR unit, a GPS receiver, smartphone sensors, vehicle CAN bus data logger and monocular, infrared are used in capturing the dataset. Although the dataset is focused on vision-based navigation, the wheel speed information from the CAN bus of the vehicle as well as the vehicle's GPS positional coordinates were used in this study. The CUPAC dataset was also collected using a Ford Fiesta at a sample rate of 10 Hz. More details of the CUPAC dataset can be found in [21].

3.2 Performance Evaluation Metrics

The following metrics as adopted in [15] were used to evaluate the performance of the WhONet, the physics model as presented in section 2.2, and the approach investigated in [19].

**Cumulative Root Square Error (CRSE):** The CRSE describes the cumulative root mean squared of the prediction error for every one second of the total duration of the GNSS outage. This method ignores the negative sign of the estimated errors thus providing a better understanding of the performance of the positioning techniques being analysed. The CRSE is defined mathematically in Equation (16).

$$\text{CRSE} = \sum_{t=1}^{N_t} \sqrt{{e_{pred}}^2} \tag{16}$$

Where $N_t$ is GNSS outage length of 10s in the challenging scenarios and 30s, 60s, 120s and 180s in the longer GNSS outage scenarios, $t$ is the sampling period and $e_{pred}$ is the prediction error.

**Cumulative True Error (CTE):** The CTE measures the summation of the prediction error from every one second time interval of the total GNSS outage duration. Since no root squared value is taken in this method, the positive and negative signs of the error estimations are considered in the calculation. Over estimations and under estimations of the position errors during the GNSS outage can be better understood with this metric. The CTE is less realistic in comparing the performances of positioning techniques compared to the CRSE. Equation (17) describes the CTE method.

$$\text{CTE} = \sum_{t=1}^{N_t} e_{pred} \tag{17}$$

**Mean (μ):** Mean refers to the statistical mean of the CRSE and CTE across all test sequences in each scenario investigated.

$$\mu_{CRSE} = \frac{1}{N_s}\sum_{i=1}^{N_s} CRSE, \mu_{CTE} = \frac{1}{N_s}\sum_{i=1}^{N_s} CTE \tag{18}$$

Where $N_s$ is the total number of test sequences in each scenario.

**Standard deviation ($\sigma$):** Standard deviation provides information on the variations of the CRSE and CTE of all test sequences in each scenario investigated.

$$\sigma_{CRSE} = \sqrt{\frac{\sum(CRSE_i - \mu_{CRSE})^2}{N_s}}, \sigma_{CTE} = \sqrt{\frac{\sum(CTE_i - \mu_{CTE})^2}{N_s}} \tag{19}$$

**Minimum:** This refers to the minimum of CRSE and CTE of all test sequences in each scenario investigated.

**Maximum:** This refers to the maximum of CRSE and CTE of all test sequences in each scenario investigated. The max metric holds more significance compared to the μ and min in the challenging driving scenarios experiment, as it captures the performance of the vehicle in each challenging scenario explored and further informs on the accuracy of the investigated techniques in each scenario.

### 3.3 Neural Network Comparative Analysis

The use of deep learning and Internet of Things on low-memory devices for complex sensing and computing applications has gained a lot of attention in recent years. However, a key impediment to their adoption on such devices remains their high computation cost. We therefore comparatively analyse the performance of the IDNN, GRU, classic RNN and LSTM on compact parameterisation for use in the proposed WhONet for accurate positional tracking. A detailed comparison of the IDNN, GRU, RNN and LSTM is presented on Table A1 in the appendix. With an analysis performed on the 180s longer GNSS outage using the CTE and CRSE metric, the result so obtained shows that the RNN is able to provide the most accurate estimations of the NNs compared while using a lower number of weighted parameters as presented on Table 2. Although the IDNN has fewer weighted parameters compared to the RNN, the RNN offers the best balance between accuracy and computational complexity. The performance of the RNN on the wheel-speed-based position tracking reveals that the dynamic information from the short-term memory of the vehicle's motion is more important than long-term memory and, as such, is less affected by the vanishing gradient issue of the RNN. As such, the RNN is selected for use in WhONet. The models are trained according to Section 3.4.

**Table 2.** Number of trainable parameters in each neural network (NN) across various weighted connections.

| Number of Neurons | Number of Trainable Parameters | | | |
|---|---|---|---|---|
| | RNN (2-Layer) | GRU (2-Layer) | LSTM (2-Layer) | IDNN (2-Layer) |
| 32 | 2,369 | 7,041 | 9,377 | 1,345 |
| 48 | 4,321 | 12,865 | 17,137 | 2,017 |
| 64 | 6,785 | 20,225 | 26,945 | 2,689 |
| 72 | 8,209 | 24,481 | 32,617 | 3,025 |
| 128 | 21,761 | 65,025 | 86,657 | 5,377 |
| 256 | 76,289 | 228,353 | 304,385 | 10,753 |
| 512 | 283,649 | 849,921 | 1,133,057 | 21,505 |

### 3.4 Training of the WhONet Models

WhONet is trained using the Keras–Tensorflow platform on the IO-VNB data subsets presented in Table A2 and the CUPAC data subsets presented in Table A3. The IO-VNB datasets used to train the WhONet Model are characterised by about 1590 mins of drive time over a total distance of 1,165 km, whilst the CUPAC trainset is characterised by about 104 mins of drive over 54.9 km. The trained model on the IO-VNB dataset is used to evaluate the challenging driving scenarios in section 4.1 and the IO-VNBD longer-term GNSS outage scenarios discussed in section 4.2. The model trained on the CUPAC trainsets are used to evaluate the CUPAC longer-term GNSS outages also discussed in section 4.2. The models were optimised using the adamax optimiser with an initial learning rate of 0.0007 and trained using the mean absolute error loss function and a dropout rate of 0.05 of the weighted connections in the hidden layers. Furthermore, the input features to the models were normalised between 0 and 1 to reduce learning bias. Table 3 highlights the parameters characterising the training of the WhONet model.

**Table 3.** WhONet's training parameters.

| Parameters | Displacement Estimation |
|---|---|
| Learning rate | 0.0007 |
| Dropout rate | 0.05 |
| Time step | 1 |
| Hidden layers | 1 |
| Hidden neurons | 72 |
| Batch size | 128 |

3.5 WhONet's Evaluation

The data subsets used in investigating the performance of the WhONet models and the physics model are presented in Tables A4, A5, A6 and A7. Although eventually evaluated on the complex scenarios, as mentioned previously, the performance of the WhONet is first examined on a relatively easy scenario, i.e., an approximate straight-line travel on the motorway. The evaluation of the models on the motorway scenario helps to gauge their performance on a relatively easy driving situation before they are tested on more complex scenarios and longer GNSS scenarios of 30s, 60s, 120s and 180s. Nonetheless, the motorway scenario could be considered challenging due to the large distance covered per second. GPS outages are assumed on the test scenarios, for the purpose of the investigation with a prediction frequency of 1s.

Each test set used in the challenging (complex) scenario is divided into several test sequences of 10 seconds length each. However, in the longer term GNSS outage scenarios, the datasets are broken down into test sequences of 30s, 60s, 120s or 180s depending on the outage scenario being evaluated. The maximum, minimum, average and standard deviation, of the CRSE's and CTE's of all the test sequences evaluated within each dataset are recorded in each scenario and used in evaluating the performance of each model.

## 4. Results and Discussion

In this section, the performance of the WhONet is evaluated in comparison to the results obtained from the physics model. The performance evaluation is first performed on the challenging driving scenarios described in section 2.1, and then the longer-term GNSS outage scenarios described in section 2.2 using the metrics defined in section 3.2.

4.1 Challenging Scenarios

The performance of the WhONet is first evaluated on the motorway scenario using the V-Vw12 dataset as in [19]. The motorway scenario presents a less challenging scenario with vehicular travel on an approximate straight-line. More analysis is then done on more challenging scenarios such as hard-brake, roundabouts, quick changes in acceleration and sharp cornering and successive left and right turns. In this sub-section, the performance of the WhONet is also compared to the investigations performed in [19].

4.1.1 Motorway Scenario

As reported on Table 4, WhONet outperforms the approach proposed in [19] and the physics model by a corresponding 57% and 83% error reduction on the max CTE metric and a corresponding 52% and 83% error reduction on the max CRSE metric over an average travel distance of 250 m. With the lowest standard deviation of 0.04, the result also shows that the proposed method is able to consistently track the motion of the vehicle over the 10 test sequences evaluated. Figure 4 further highlights the improvement the WhONet makes over the other approaches evaluated.

***Table 4.*** *Motorway GNSS outage experiment results on the IO-VNB dataset*

| **IO-VNB Dataset** | **Performance Metrics** | **Position Estimation Error (m)** | | | | | | **Total Distance Travelled (m)** | **Number of Test Sequences evaluated** |
|---|---|---|---|---|---|---|---|---|---|
| | | **CTE** | | | **CRSE** | | | | |
| | | **Physics Model** | **WhO Net** | **[19]** | **Physics Model** | **WhO Net** | **[19]** | | |
| V_w12 | Max | **1.34** | **0.23** | **0.53** | **1.96** | **0.33** | **0.68** | 268 | 10 |
| | Min | 0.04 | 0.00 | 0.46 | 0.36 | 0.13 | 0.25 | 235 | |
| | µ | 0.52 | 0.10 | 0.32 | 1.07 | 0.26 | 0.46 | 250 | |
| | $\sigma$ | 0.46 | 0.08 | 0.21 | 0.45 | 0.04 | 0.08 | 11 | |

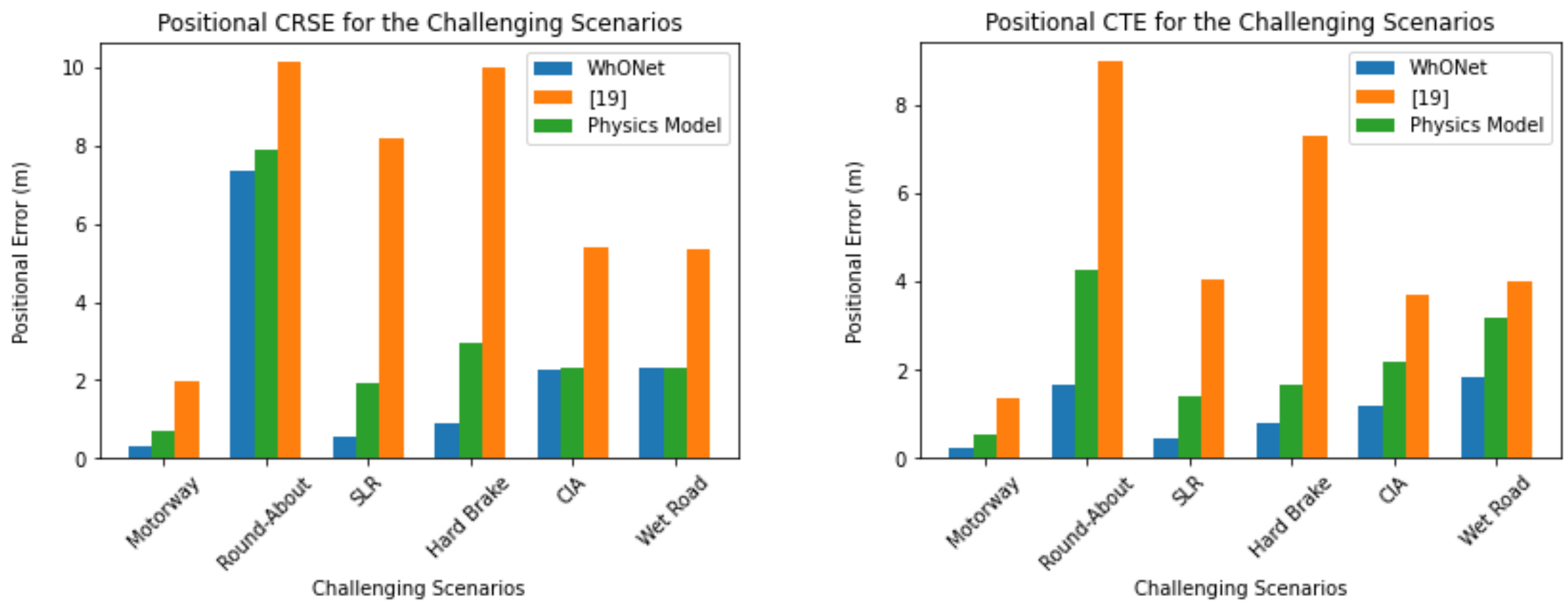

***Figure 4.*** *Comparison of the positional max CRSE and max CTE of all models evaluated on the challenging scenarios.*

#### 4.1.2 Roundabout Scenario

Two datasets were used in the Roundabout analysis as reported on Table 5. In both datasets the standard deviation of the WhONet model is found to be the minimum of all approaches compared. With consistent low standard deviations, it is evident that the proposed method is capable of tracking the vehicle's position consistently. On the max CTE metrics, the WhONet approach shows up to a 85% and 97% reduction of the error estimations compared to the approach in [19] and that obtained from the physics model, respectively. However, on the max CRSE metrics, the corresponding percentage reductions in the errors are up to 68% and 94%, compared to the approach in [19] and that obtained from the physics model over a maximum travel distance of 262 m which features a minimum of one roundabout. The roundabout proves to be the most challenging scenario for the proposed method just as in other studies done in [15] and [19].

***Table 5.*** *Roundabout challenging GNSS outage experiment results on the IO-VNB dataset*

| IO-VNB Dataset | Performance Metrics | Position Estimation Error (m) CTE | | | CRSE | | | Total Distance Travelled (m) | Number of Test Sequences evaluated |
|---|---|---|---|---|---|---|---|---|---|
| | | Physics Model | WhO Net | [19] | Physics Model | WhO Net | [19] | | |
| V_Vta11 | Max | **8.98** | **1.66** | **4.28** | **10.13** | **7.35** | **7.88** | **262** | 5 |
| | Min | 0.47 | 0.00 | 0.15 | 1.50 | 0.11 | 0.24 | 85 | |
| | μ | 3.36 | 0.56 | 1.48 | 5.65 | 1.75 | 2.42 | 159 | |
| | σ | 3.22 | 0.48 | 1.48 | 3.74 | 2.66 | 2.78 | 67 | |
| V_Vfb02d | Max | **7.57** | **0.22** | **1.45** | **7.57** | **0.49** | **1.55** | **151** | 8 |
| | Min | 0.26 | 0.00 | 0.15 | 0.98 | 0.14 | 0.37 | 14 | |
| | μ | 1.98 | 0.11 | 0.73 | 3.47 | 0.30 | 1.04 | 96 | |
| | σ | 2.34 | 0.07 | 0.46 | 2.05 | 0.09 | 0.37 | 52 | |

#### 4.1.3 Quick Changes in Vehicles Acceleration Scenario

Two datasets are analysed in the quick changes in acceleration scenario. As with the previous scenarios, the WhONet method produced the lowest max CTE and CRSE. As presented on Table 6, the WhONet method shows up to a 64% reduction on the max CTE estimations compared to that obtained from the physics model and up to 45% against that proposed in [19] However, on the max CRSE metric, the percentage error reductions are up to 58% against the physics model and up to 5% against [19]. Furthermore, the standard deviation of the estimations is lowest with the WhONet compared to the other approaches with values of 0.38 and 0.49 on the CTE and CRSE metrics, respectively.

***Table 6.*** *Showing the quick changes in vehicular acceleration challenging GNSS outage experiment results on the IO-VNB dataset*

| IO-VNB Dataset | Performance Metrics | Position Estimation Error (m) | | | | | | Total Distance Travelled (m) | Number of Test Sequences evaluated |
|---|---|---|---|---|---|---|---|---|---|
| | | CTE | | | CRSE | | | | |
| | | Physics Model | WhO Net | [19] | Physics Model | WhO Net | [19] | | |
| V_Vfb02e | Max | **3.35** | **1.20** | **2.17** | **4.64** | **2.82** | **2.95** | **192** | 9 |
| | Min | 0.08 | 0.00 | 0.15 | 1.49 | 0.17 | 0.54 | 123 | |
| | μ | 1.65 | 0.53 | 0.92 | 2.94 | 1.14 | 1.43 | 156 | |
| | $\sigma$ | 1.17 | 0.39 | 0.60 | 1.01 | 0.81 | 0.65 | 19 | |
| V_Vta12 | Max | **3.72** | **1.62** | **1.85** | **5.39** | **2.27** | **2.30** | **218** | 6 |
| | Min | 0.20 | 0.00 | 0.23 | 2.16 | 0.14 | 0.65 | 145 | |
| | μ | 1.83 | 0.63 | 0.74 | 3.76 | 0.89 | 1.37 | 178 | |
| | $\sigma$ | 1.03 | 0.38 | 0.55 | 1.13 | 0.49 | 0.55 | 24 | |

#### 4.1.4 Hard Brake Scenario

In the hard brake scenario, it can be observed that compared to the approach in [19] and the estimations from the physics model, the WhONet continues to consistently track the motion of the vehicle with lowest error estimations on all metrics considered. In both datasets considered, the WhONet approach was able to estimate the error with less than 1 m accuracy on both the CTE and CRSE metrics. When comparing the performance of WhONet on the V_Vw16b dataset (see Table A5) to that proposed in [19] and the errors obtained from the physics model, it can be observed that there is a 70% and 91% respective reduction of the error estimations on the max CRSE metrics. On the V_Vw17 dataset (see Table A5), the max CRSE reductions are reported to be in the order of 59% compared to [19] and 89% compared to the physics model as presented on Table 7.

***Table 7.*** *Hard Brake challenging GNSS outage experiment results on the IO-VNB dataset*

| IO-VNB Dataset | Performance Metrics | Position Estimation Error (m) | | | | | | Total Distance Travelled (m) | Number of Test Sequences evaluated |
|---|---|---|---|---|---|---|---|---|---|
| | | CTE | | | CRSE | | | | |
| | | Physics Model | WhO Net | [19] | Physics Model | WhO Net | [19] | | |
| V_Vw16b | Max | **7.31** | **0.81** | **1.65** | **9.99** | **0.88** | **2.93** | **234** | 11 |
| | Min | 0.00 | 0.00 | 0.02 | 0.80 | 0.10 | 0.10 | 87 | |
| | μ | 2.30 | 0.20 | 0.74 | 4.04 | 0.36 | 1.49 | 170 | |
| | $\sigma$ | 2.44 | 0.23 | 0.46 | 3.04 | 0.22 | 0.95 | 50 | |
| V_Vw17 | Max | **2.91** | **0.37** | **1.69** | **8.07** | **0.87** | **2.13** | **175** | 3 |
| | Min | 0.80 | 0.20 | 1.21 | 5.52 | 0.61 | 3.41 | 157 | |
| | μ | 1.94 | 0.26 | 1.45 | 6.40 | 0.78 | 3.51 | 169 | |
| | $\sigma$ | 0.87 | 0.01 | 0.24 | 1.19 | 0.01 | 0.01 | 8 | |

#### 4.1.5 Sharp Cornering and Successive Left-Right Turns Scenario (SLR)

Table 8 presents the results from the sharp cornering and successive left and right turns scenario. From observation, the WhONet method outperforms the other two approaches considered. It can also be seen that the WhONet remains consistent in accurately estimating the positioning error. On comparing the performance of WhONet to [19] and the error estimations from the physics model, the WhONet offers a corresponding error reduction of up to 70% and up to 93% on the max CRSE metric.

***Table 8.*** *Successive left and right turns and sharp cornering challenging GNSS outage experiment results on the IO-VNB dataset*

| IO-VNB Dataset | Performance Metrics | Position Estimation Error (m) | | | | | | Total Distance Travelled (m) | Number of Test Sequences evaluated |
|---|---|---|---|---|---|---|---|---|---|
| | | CTE | | | CRSE | | | | |
| | | Physics Model | WhO Net | [19] | Physics Model | WhO Net | [19] | | |
| V_Vw6 | Max | **3.68** | **0.35** | **0.73** | **5.00** | **0.57** | **1.45** | **107** | 12 |
| | Min | 0.09 | 0.00 | 0.06 | 0.73 | 0.11 | 0.34 | 60 | |
| | µ | 0.97 | 0.14 | 0.36 | 2.75 | 0.36 | 0.84 | 86 | |
| | $\sigma$ | 1.15 | 0.11 | 0.20 | 1.42 | 0.09 | 0.29 | 12 | |
| V_Vw7 | Max | **5.08** | **0.50** | **1.36** | **6.90** | **0.61** | **1.93** | **115** | 15 |
| | Min | 0.03 | 0.00 | 0.10 | 1.50 | 0.08 | 0.29 | 16 | |
| | µ | 2.06 | 0.21 | 0.54 | 3.87 | 0.36 | 1.14 | 70 | |
| | $\sigma$ | 1.40 | 0.14 | 0.39 | 1.33 | 0.13 | 0.46 | 25 | |
| V_Vw8 | Max | **4.05** | **0.46** | **1.42** | **8.19** | **0.57** | **1.91** | **109** | 16 |
| | Min | 0.05 | 0.00 | 0.05 | 1.13 | 0.11 | 0.43 | 36 | |
| | µ | 1.36 | 0.17 | 0.57 | 4.09 | 0.37 | 1.19 | 73 | |
| | $\sigma$ | 1.30 | 0.12 | 0.38 | 2.04 | 0.12 | 0.44 | 20 | |

#### 4.1.6 Wet Road Scenario

Wet road conditions are one of the most critical driving scenarios that challenges an accurate tracking of wheeled vehicles due to wheel drifts, particularly on wet and muddy roads. However, the results presented on Table 9 show that the WhONet is able to capture the uncertainties caused by the wheel slippage. As reported on Table 9, it can be observed that the standard deviations of the error estimations of WhONet on datasets V_Vtb8 and V_Vtb11 are in the range of 0.01- 0.02 on both the CTE and CRSE metrics, making the WhONet a reliable method for tracking in such scenarios. On the max CRSE and CTE metric, the WhONet achieves a percentage error reduction of up to 62% and 92%, respectively compared to [19], and up to 94% and 97%, respectively, compared to that obtained from the physics model.

However, the WhONet reports a similar performance to [19] on the V_Vtb13 dataset (see Table A5) using the max CRSE metric.

***Table 9.*** *Wet road challenging GNSS outage experiment results on the IO-VNB dataset*

| **IO-VNB Dataset** | **Performance Metrics** | **Position Estimation Error (m)** | | | | | | **Total Distance Travelled (m)** | **Number of Test Sequences evaluated** |
|---|---|---|---|---|---|---|---|---|---|
| | | **CTE** | | | **CRSE** | | | | |
| | | **Physics Model** | **WhO Net** | **[19]** | **Physics Model** | **WhO Net** | **[19]** | | |
| V_Vtb8 | Max | **2.21** | **0.07** | **0.89** | **2.21** | **0.19** | **0.48** | **202.8** | 6 |
| | Min | 0.02 | 0.00 | 0.10 | 1.28 | 0.08 | 0.23 | 180.4 | |
| | µ | 1.17 | 0.03 | 0.52 | 1.66 | 0.14 | 0.33 | 192.7 | |
| | $\sigma$ | 0.69 | 0.02 | 0.27 | 0.39 | 0.02 | 0.04 | 8.4 | |
| V_Vtb11 | Max | **1.81** | **0.06** | **0.38** | **2.28** | **0.14** | **0.37** | **199.7** | 4 |
| | Min | 0.55 | 0.00 | 0.18 | 1.03 | 0.05 | 0.24 | 191.3 | |
| | µ | 1.07 | 0.04 | 0.31 | 1.63 | 0.11 | 0.31 | 194.5 | |
| | $\sigma$ | 0.46 | 0.01 | 0.09 | 0.49 | 0.02 | 0.03 | 3.4 | |
| V_Vtb13 | Max | **4.01** | **1.82** | **3.20** | **5.36** | **2.33** | **2.33** | **114.9** | 12 |
| | Min | 0.03 | 0.00 | 0.11 | 1.85 | 0.06 | 0.20 | 46.3 | |
| | µ | 1.28 | 0.59 | 0.78 | 3.53 | 0.75 | 1.24 | 79.2 | |
| | $\sigma$ | 1.22 | 0.39 | 0.87 | 1.17 | 0.49 | 0.73 | 24.2 | |

### 4.2 Longer-term GNSS outage

WhONet's performance is also evaluated on four longer-term GNSS outages of 30 seconds, 60 seconds, 120 seconds and 180 seconds using the datasets presented on Table A4-A7. The performance evaluation is done on both the IO-VNBD dataset and the CUPAC dataset. More details on the results obtained from each longer-term GNSS outage can be found in Tables A8-A15.

#### 4.2.1 30s GNSS Outage Scenario

In the 30 seconds GNSS outage scenario, WhONet provides an error reduction on the physics model's estimation by up to 92%. As presented on Tables A8 and A9, the WhONet achieves the best average CTE and CRSE of 0.23 m and 0.67 m compared to 0.84 m and 2.31 m of the physics model over a max distance per test sequence of 987 m of all 688 sequences evaluated on the IO-VNB dataset (see Table A6). On the CUPAC dataset (see Table A7), the WhONet also obtains the best average CTE and CRSE of 0.34 m and 0.94 m compared to 1.88 m and 7.13 m of the physics model over a max distance per test sequence of 629 m of all 121 sequences evaluated, as shown on Table A9. The robustness of the WhONet is further emphasised by the low standard deviation of 0.18 obtained compared to 1.14 of the physics model as reported by both Tables A8 and A9. The result so obtained shows that the proposed model is able to significantly improve the accuracy of wheel odometry in the 30s outage scenario as shown in Figures 5 and 6.

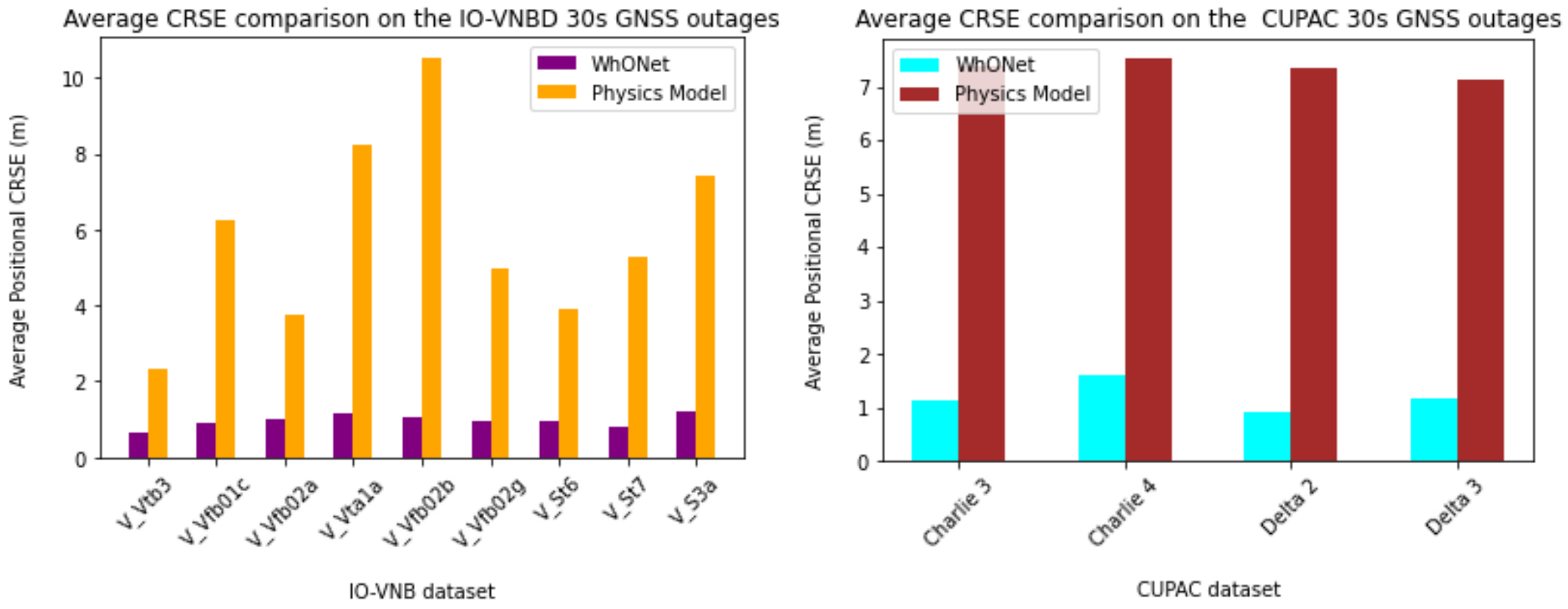

***Figure 5.*** *Comparison of the average positional CRSE on the IO-VNB and CUPAC dataset during the 30s GNSS signal outage experiment*

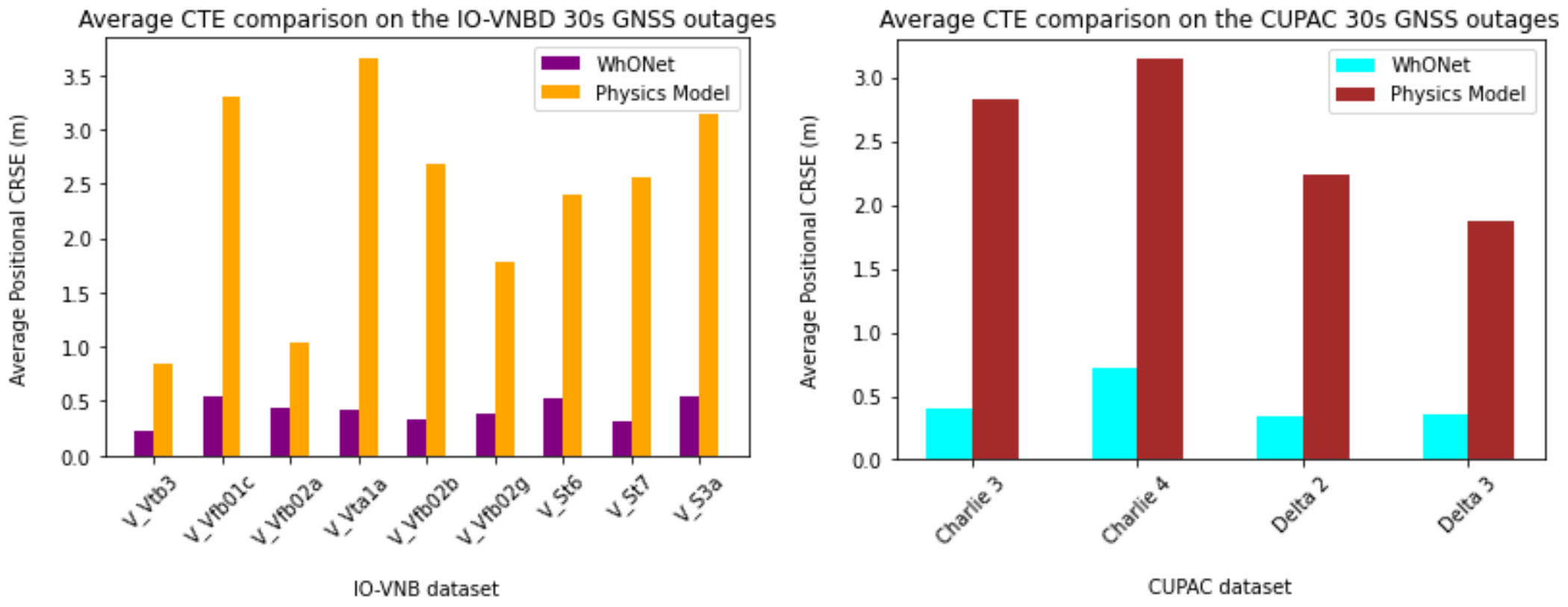

***Figure 6.*** *Comparison of the average positional CTE on the IO-VNB and CUPAC dataset during the 30s GNSS signal outage experiment*

#### 4.2.2 60s GNSS Outage Scenario

Table A10 and A11 report the results from the 60 seconds GNSS outage experiment on both the IO-VNBD and CUPAC dataset. On the CTE and CRSE metric, the WhONet provides the best average of 0.30 m and 1.31 m compared to 1.02 m and 4.56 m of the physics model over a max test sequence distance covered of 1960 m across all 342 sequences evaluated on the IO-VNBD (see Table A6). The performance evaluation on the CUPAC dataset, as presented on Table A11, reveals the best average of 0.49 m and 1.88 m compared to 2.53 m and 14.25 m of the physics model over 57 test sequences evaluated with a max displacement of 1238 m per sequence. Overall, the results show the WhONet is able to provide an error reduction of up to 91% on the physics model's estimation in the 60 s outage scenario as illustrated in Figures 7 and 8.

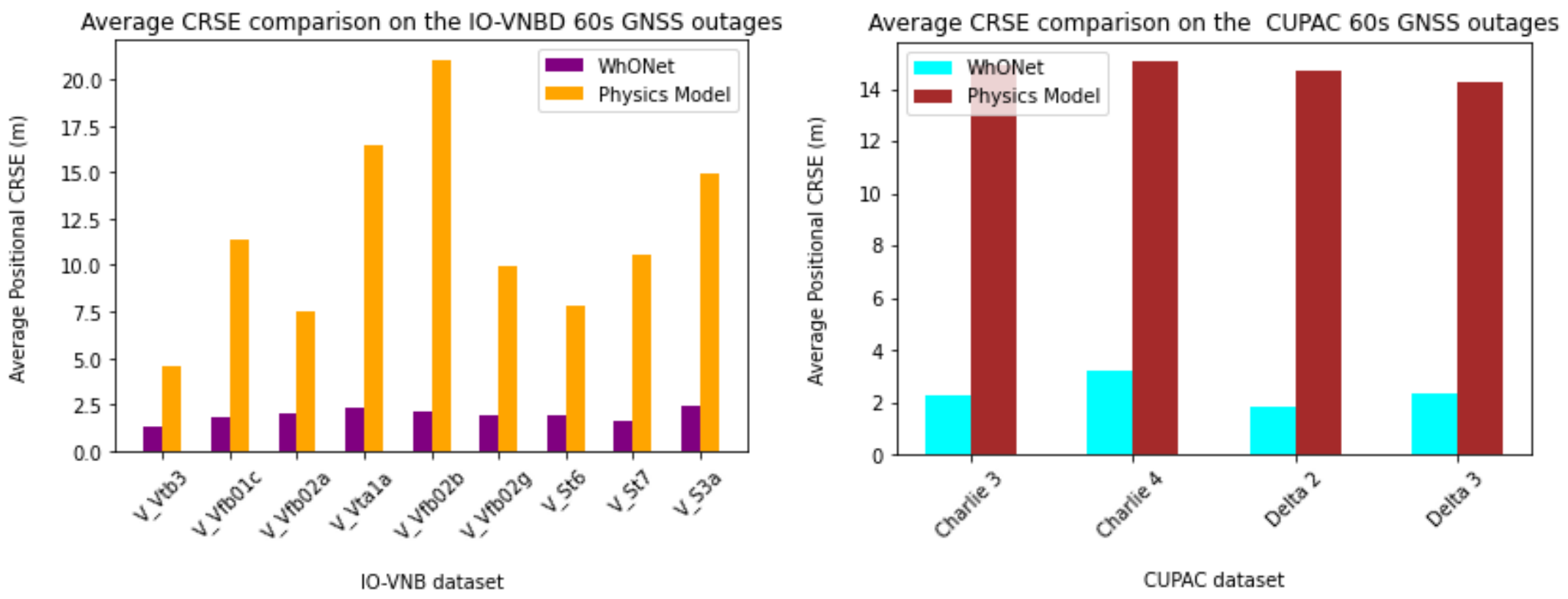

***Figure 7.*** *Comparison of the average positional CRSE on the IO-VNB and CUPAC dataset during the 60s GNSS signal outage experiment*

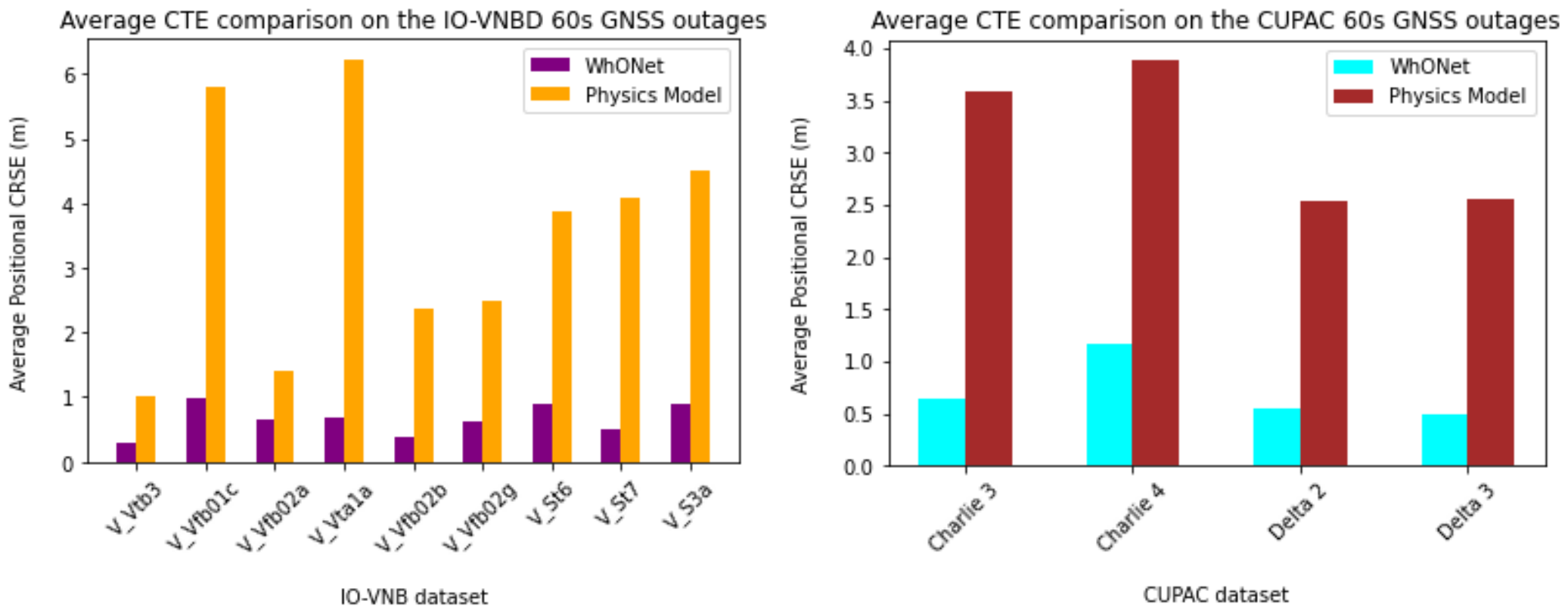

***Figure 8.*** *Comparison of the average positional CTE on the IO-VNB and CUPAC dataset during the 60s GNSS signal outage experiment*

#### 4.2.3 120s GNSS Outage Scenario

The WhONet outperforms the physics model by up to 91% on the 120 seconds GNSS outage scenario as illustrated in Figures 9 and 10. When evaluated on the IO-VNBD, as shown on Table A12, the WhONet achieves the best average CTE and CRSE of 1.78 m and 2.62 m compared to 1.78 m and 9.11 m of the physics model over a max distance per test sequence of 3.9 km of all 168 sequences. Furthermore, on the CUPAC dataset (see table A7), the WhONet also obtains the best average CTE and CRSE of 0.46 m and 3.75 m compared to 3.73 m and 28.49 m of the physics model over a max distance per test sequence of 2.3 km out of all 29 sequences evaluated as shown on Table A13. The robustness of WhONet further emphasised by the low standard deviation of 0.06 obtained compared to 0.66 of the physics model as reported by both Tables A12 and A13. The result so obtained shows that the proposed model is able to significantly improve the accuracy of wheel odometry in the 120s outage scenario.

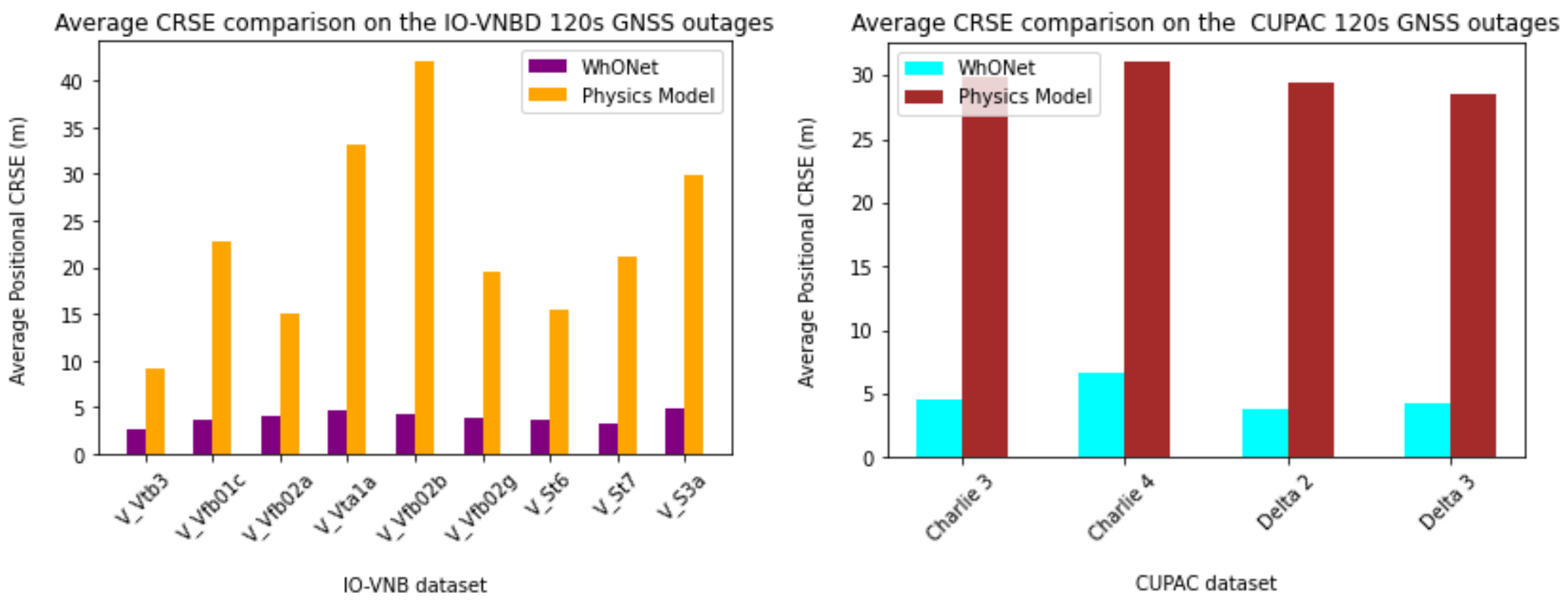

***Figure 9.*** *Comparison of the average positional CRSE on the IO-VNB and CUPAC dataset during the 120s GNSS signal outage experiment*

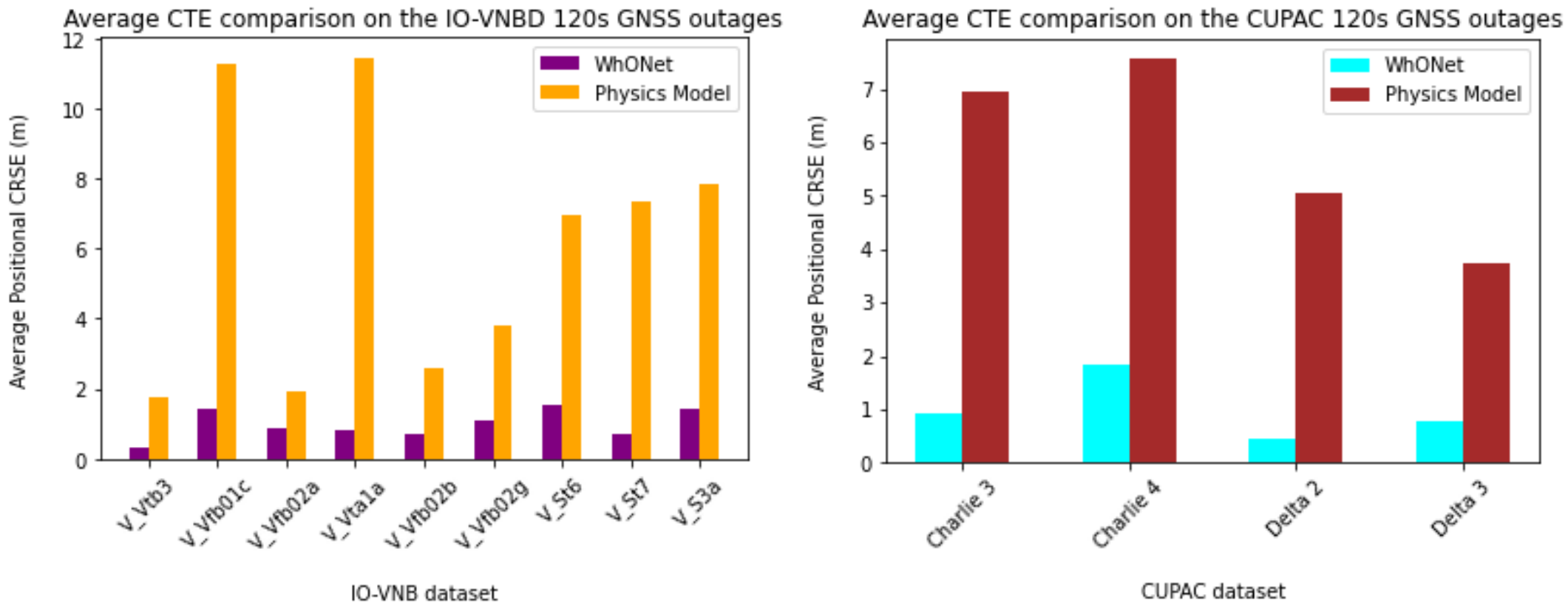

***Figure 10.*** *Comparison of the average positional CTE on the IO-VNB and CUPAC dataset during the 120s GNSS signal outage experiment*

#### 4.2.4 180s GNSS Outage Scenario

In the 180 seconds GNSS outage scenario, WhONet provides an error reduction on the physics model's estimation by up to 90%. As presented on Tables A14 and A15, the WhONet achieves the best average CTE and CRSE of 0.49 m and 3.93 m compared to 2.90 m and 13.67 m of the physics model over a max distance per test sequence of 5.6 km of all 111 sequences evaluated on the IO-VNB dataset (see Table A6). On the CUPAC dataset (see Table A7), the WhONet also obtains the best average CTE and CRSE of 0.51 m and 5.13 m compared to 4.59 m and 42.74 m of the physics model over a max distance per test sequence of 3.4 km of all 18 sequences evaluated as shown on Table A15. The robustness of the WhONet is further emphasised by the low standard deviation of 0.01 obtained compared to 1.25 of the physics model as reported by both Tables A14 and A15. The result obtained shows that the proposed model is able to significantly improve the performance of wheel odometry in the 180 s outage scenario. The WhONet's improvement over the physics model on each dataset evaluated on the average CTE and CRSE is presented on Figures 11, 12 and 13. Figure 14 shows a sample trajectory of the vehicle during the longer-term outage scenario.

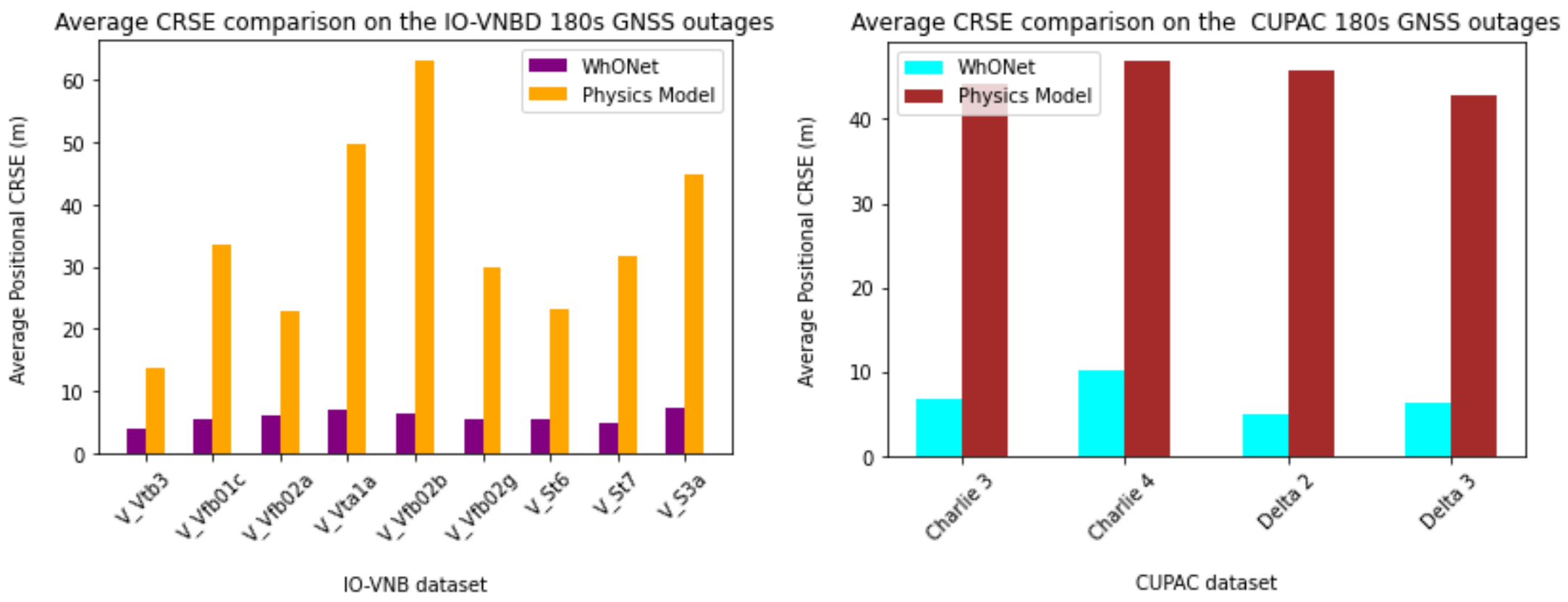

***Figure 11.*** *Comparison of the average positional CRSE on the IO-VNB and CUPAC dataset during the 180s GNSS signal outage experiment*

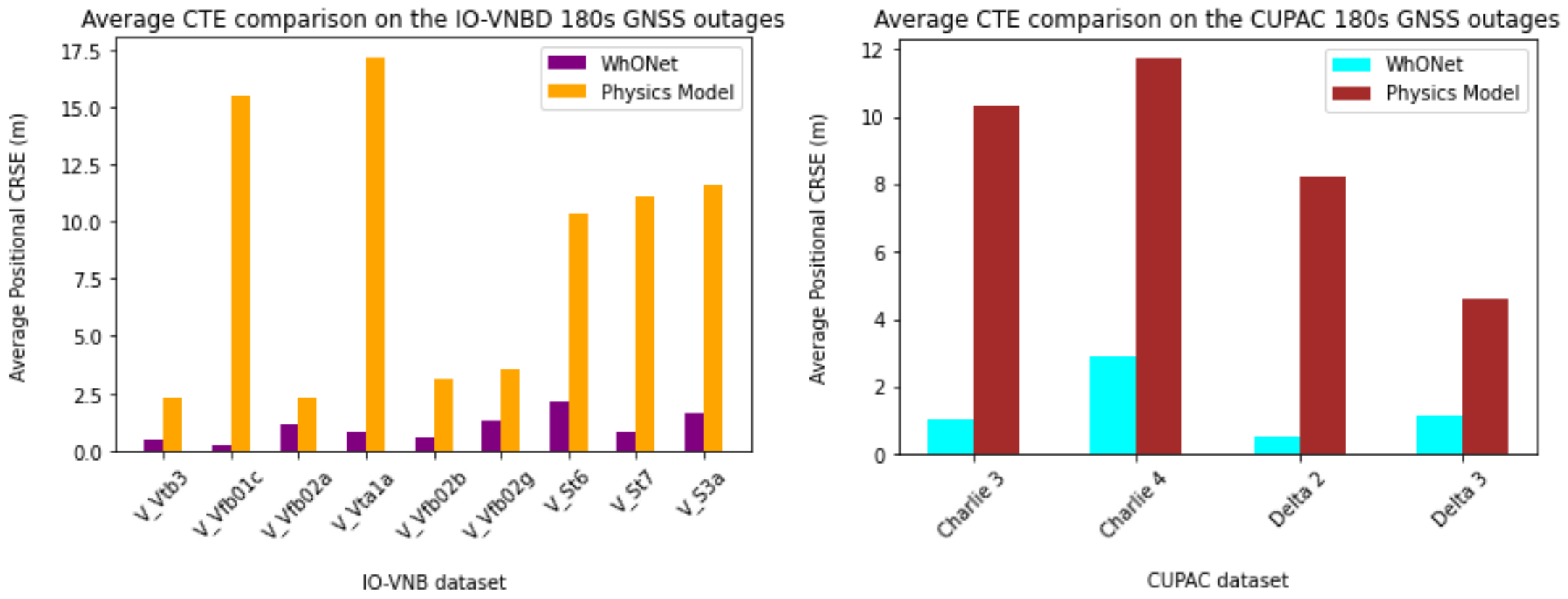

***Figure 12.*** *Comparison of the average positional CTE on the IO-VNB and CUPAC dataset during the 180s GNSS signal outage experiment*

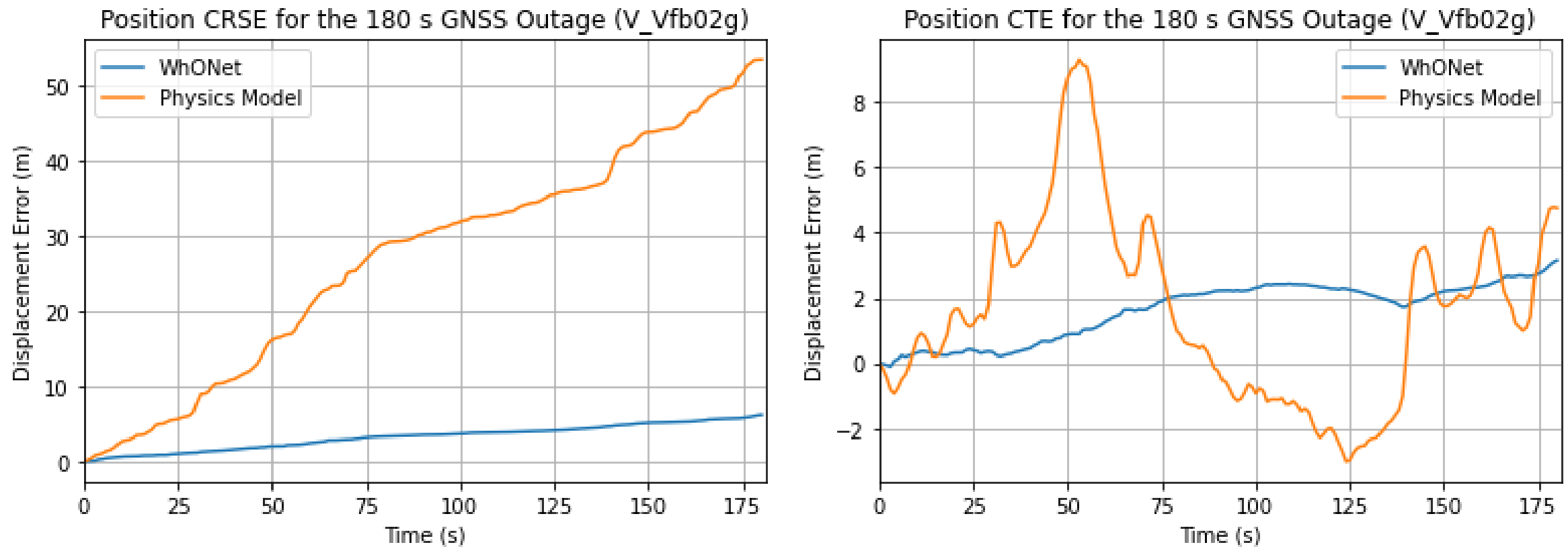

***Figure 13.*** *A sample evolution of the CRSE and CTE in the 180 s GNSS outage scenario*

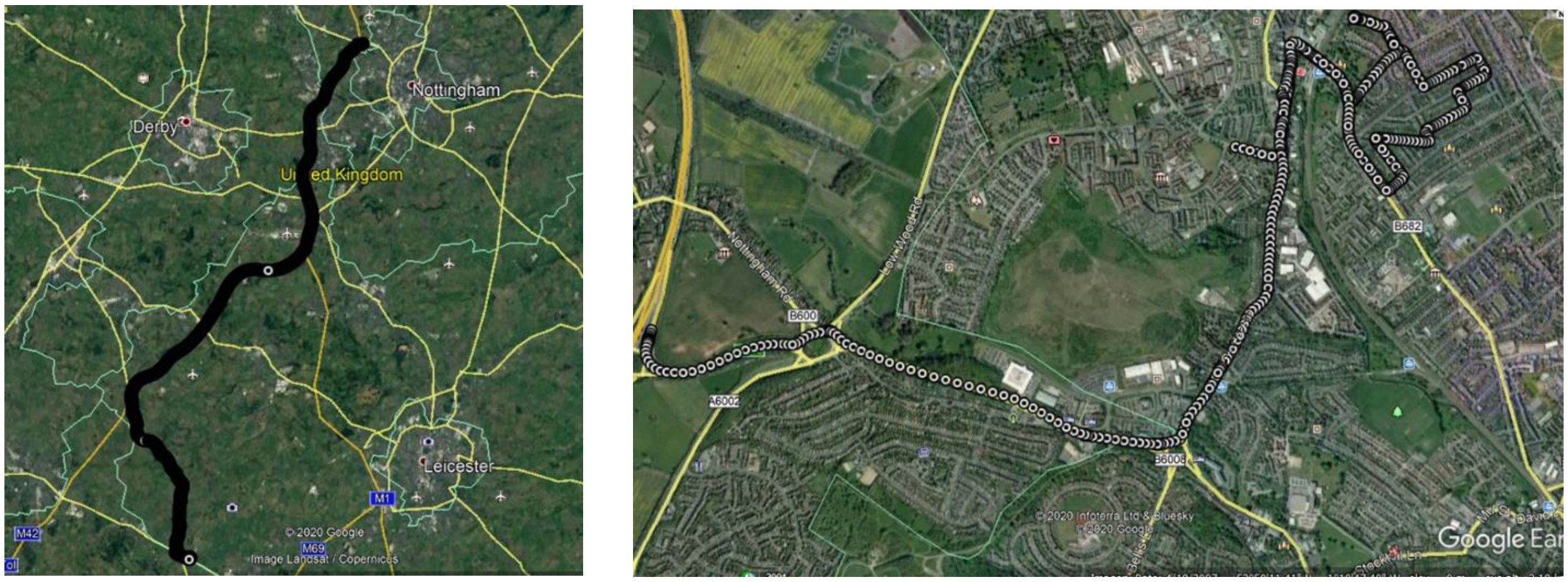

***Figure 14.*** *The vehicle trajectory on the V-Vfb02g dataset (left) and V-Vfbo2b dataset (right) used in evaluating WhONet's performance during the 30s, 60s, 120s and 180s GNSS outage experiment.*

## 5. Conclusions

Safe navigation of autonomous vehicles in GNSS-deprived environments is dependent on reliable and robust vehicle positioning techniques. The availability of GNSS for road localisation is affected in urban canyons, mountains, tunnels, under bridges, under dense tree canopies, etc. Wheel odometry information can be used to continuously position the vehicle in the absence of GNSS signals. In this paper, we proposed a deep learning approach called WhONet for the positional tracking of wheeled vehicles and other robots alike. We evaluated the performance of the proposed approach on several challenging vehicular driving scenarios such as hard brake, roundabout, wet roads, sharp cornering, successive left and right turns, etc. The performance of the proposed WhONet was extensively evaluated on longer GNSS outage scenarios comprising of 809 test sequences of 30 s simulated GNSS outage, 399 sequences of 60 s GNSS outage, 197 sequences of 120 s GNSS outage and 129 sequences of 180 s GNSS outage. We demonstrated that the WhONet is able to provide up to 93% error reduction on the directly integrated positioning information from the wheel encoder. The WhONet is able to provide an accuracy of 8.62 m after 5.6 km of travel.Further research will involve exploring techniques to improve the generalisability of WhONet to other vehicle types and journey domains to improve the accuracy and robustness of its positional output.

## 6. Appendix

**Table A1.** Performance comparison of the IDNN, LSTM, GRU and RNN using the CTE and CRSE metric.

| Dataset | Performance Metrics | CTE (m) | | | | CRSE (m) | | | |
|---|---|---|---|---|---|---|---|---|---|
| | | IDNN | LSTM | GRU | RNN | IDNN | LSTM | GRU | RNN |
| V_Vtb3 | Max | 1.15 | 1.13 | 1.12 | 1.06 | 5.42 | 6.20 | 5.23 | 5.29 |
| | Min | 0.00 | 0.00 | 0.00 | 0.00 | 2.99 | 4.02 | 2.90 | 3.06 |
| | µ | **0.44** | 0.63 | 0.53 | 0.55 | **4.20** | 4.97 | 4.23 | 4.22 |
| | *σ* | 0.19 | 0.17 | 0.22 | 0.17 | 0.24 | 0.51 | 0.26 | 0.19 |
| V_Vfb01c | Max | 7.66 | 6.20 | 5.52 | 4.99 | 10.19 | 10.41 | 9.53 | 9.16 |
| | Min | 0.01 | 0.01 | 0.03 | 0.01 | 4.89 | 6.37 | 4.88 | 4.74 |
| | µ | 3.68 | 3.02 | 3.22 | **3.00** | 7.06 | 8.03 | 7.06 | **6.64** |
| | *σ* | 2.19 | 1.29 | 0.68 | 0.29 | 1.62 | 1.42 | 1.39 | 1.33 |
| V_Vfb02a | Max | 9.05 | 9.45 | 8.43 | 5.80 | 14.74 | 15.29 | 14.15 | 13.41 |
| | Min | 0.00 | 0.00 | 0.00 | 0.00 | 4.36 | 4.14 | 4.59 | 3.77 |
| | µ | 3.99 | 3.30 | 3.65 | **2.80** | 7.79 | 7.90 | 8.01 | **7.19** |
| | *σ* | 2.33 | 2.50 | 2.12 | 1.45 | 2.02 | 2.48 | 1.94 | 1.82 |
| V_Vta1a | Max | 5.86 | 6.13 | 4.63 | 4.28 | 13.84 | 14.99 | 13.66 | 13.68 |
| | Min | 0.00 | 0.00 | 0.00 | 0.00 | 5.19 | 6.02 | 5.09 | 5.44 |
| | µ | 3.07 | 3.24 | 2.10 | **2.07** | 8.18 | 9.95 | 8.11 | **7.71** |
| | σ | 1.04 | 1.44 | 1.07 | 0.94 | 1.99 | 2.27 | 2.06 | 2.08 |
| V_Vfb02b | Max | 1.65 | 2.94 | 1.54 | 1.34 | 9.87 | 14.99 | 10.33 | 9.70 |
| | Min | 0.00 | 0.00 | 0.01 | 0.01 | 4.63 | 7.61 | 4.74 | 4.55 |
| | µ | 0.87 | 1.53 | 0.95 | **0.76** | 7.62 | 11.05 | 7.50 | **7.29** |
| | σ | 0.41 | 0.82 | 0.45 | 0.42 | 1.53 | 2.23 | 1.39 | 1.28 |
| V_Vfb02g | Max | 7.44 | 10.62 | 7.68 | 6.94 | 10.09 | 14.25 | 10.39 | 9.57 |
| | Min | 0.05 | 0.01 | 0.00 | 0.00 | 5.60 | 6.80 | 5.49 | 5.03 |
| | µ | 5.31 | 4.74 | 4.98 | **3.76** | 8.21 | 9.31 | 8.15 | **7.24** |
| | σ | 1.97 | 1.58 | 2.16 | 1.84 | 1.21 | 1.73 | 1.03 | 1.11 |
| V_St6 | Max | 9.72 | 5.03 | 9.21 | 9.12 | 11.11 | 10.95 | 12.11 | 10.78 |
| | Min | 0.00 | 0.00 | 0.00 | 0.00 | 3.49 | 3.91 | 3.25 | 3.46 |
| | µ | 2.72 | **2.13** | 2.93 | 2.78 | 6.40 | **6.13** | 6.54 | 6.29 |
| | σ | 2.29 | 1.17 | 2.65 | 2.32 | 1.62 | 1.44 | 1.97 | 1.69 |
| V_St7 | Max | 4.30 | 4.47 | 4.01 | 4.15 | 9.60 | 10.98 | 9.65 | 9.70 |
| | Min | 0.00 | 0.00 | 0.00 | 0.00 | 2.85 | 3.45 | 2.92 | 2.71 |
| | µ | 1.85 | 1.69 | **1.60** | 1.61 | 5.74 | 6.96 | **5.58** | 5.70 |
| | σ | 0.81 | 1.32 | 0.85 | 0.78 | 1.49 | 2.28 | 1.72 | 1.61 |
| V_S3a | Max | 5.38 | 4.54 | 5.42 | 5.14 | 11.86 | 12.30 | 12.24 | 12.00 |
| | Min | 0.00 | 0.00 | 0.00 | 0.01 | 4.79 | 5.65 | 4.76 | 4.61 |
| | µ | 2.20 | 2.67 | 2.08 | **2.06** | 7.93 | 9.35 | **7.67** | 7.85 |
| | σ | 1.34 | 1.37 | 1.23 | 1.20 | 1.60 | 1.60 | 1.60 | 1.56 |

**Table A2:** IO-VNB data subsets used in training WhONet

| IO-VNB Dataset | Features |
|---|---|
| V-S1 | B-road (B4101), roundabout (x9), reverse (x5), hilly road, A4053 (ring-road), hard-brake, tyre pressure E |
| V-S2 | B-road (B4112, B4065), roundabout (x18), reverse drive (x8), motorway, dirt road, u-turn (x5), country road, successive left-right turns, hard-brake, A-roads (A4600), tyre pressure E |
| V-S3c | Roundabout(x4), A-road (A428), country roads, tyre pressure E |
| V-S4 | Roundabout (x14), u-turn, A-road, successive left-right turns, swift maneuvers, change in speed, night-time, A-road (A429, A45, A46), ring-road (A4053), tyre pressure E |
| V-St1 | Roundabout (x9), A-road (A452), B-road, car park navigation, tyre pressure E |
| V-M | Roundabout (x30), successive left-right turns, hard-brake (x21), swift maneuvers(x5), country roads, sharp turn left/right, daytime, u-turn (x1), u-turn reverse (x7), tyre pressure E |
| V-Y2 | Roundabout(x9), u-turn/reverse(x1), A-road, B-road, country road, tyre pressure E |
| V-Vta2 | Round About (×2), A Road (A511, A5121, A444), Country Road, Hard Brake, Tyre Pressure A |
| V-Vta8 | Town Roads (Build-up), A-Roads (A511), Tyre Pressure A |
| V-Vta9 | Hard-brakes, A–road (A50), tyre pressure A |
| V-Vta10 | Round About (×1), A—Road (A50), Tyre Pressure A |
| V-Vta13 | A-road (A515), country road, hard-brakes, tyre pressure A |
| V-Vta16 | Round-About (×3), Hilly Road, Country Road, A-Road (A515), Tyre Pressure A |
| V-Vta17 | Hilly Road, Hard-Brake, Stationary (No Motion), Tyre Pressure A |
| V-Vta20 | Hilly Road, Approximate Straight-line travel, Tyre Pressure A |
| V-Vta21 | Hilly Road, Tyre Pressure A |
| V-Vta22 | Hilly Road, Hard Brake, Tyre Pressure A |
| V-Vta27 | Gravel Road, Several Hilly Road, Potholes, Country Road, A-Road (A515), Tyre Pressure A |
| V-Vta28 | Country Road, Hard Brake, Valley, A-Road (A515) |
| V-Vta29 | Hard Brake, Country Road, Hilly Road, Windy Road, Dirt Road, Wet Road, Reverse (×2), Bumps, Rain, B-Road (B5053), Country Road, U-Turn (×3), Windy Road, Valley, Tyre Pressure A |
| V-Vta30 | Rain, Wet Road, U-Turn (×2), A-Road (A53, A515), Inner Town Driving, B-Road (B5053), Tyre Pressure A |
| V-Vtb1 | Valley, rain, Wet-Road, Country Road, U-T urn (×2), Hard-Brake, Swift-Manoeuvre, A—Road (A6, A6020, A623, A515), B-Road (B6405), Round About (×3), day Time, Tyre Pressure A |
| V-Vtb2 | Country Road, Wet Road, Dirt Road, Tyre Pressure A |
| V-Vtb5 | Dirt Road, Country Road, Gravel Road, Hard Brake, Wet Road, B Road (B6405, B6012, B5056), Inner Town Driving, A-Road, Motorway (M42, M1), Rush hour(Traffic) Round-About (×6), A-Road (A5, A42, A38, A615, A6), Tyre Pressure A |
| V-Vtb9 | Approximate straight-line motion, night-time, wet road, hard-brakes, A-road (A5), tyre pressure A |
| V-Vw4 | Round-About (×77), Swift-Manoeuvres, Hard-Brake, Inner City Driving, Reverse, A-Road, Motorway (M5, M40, M42), Country Road, Successive Left-Right Turns, Daytime, U-Turn (×3), Tyre Pressure D |
| V-Vw5 | Successive Left-Right Turns, Daytime, Sharp Turn Left/Right, Tyre Pressure D |
| V-Vw14b | Motorway (M42), Night-time, Tyre Pressure D |
| V-Vw14c | Motorway (M42), Round About (×2), A-Road (A446), Night-time, Hard Brake, Tyre Pressure D |
| V-Vfa01 | A-Road (A444), Round About (×1), B–Road (B4116) Day Time, Hard Brake, Tyre Pressure A |
| V-Vfa02 | B-Road (B4116), Round About (×5), A Road (A42, A641), Motorway (M1, M62) High Rise Buildings, Hard Brake, Tyre Pressure C |
| V-Vfb01a | City Centre Driving, Round-About (×1), Wet Road, Ring Road, Night, Tyre Pressure C |
| V-Vfb01b | Motorway (M606), Round-About (×1), City Roads Traffic, Wet Road, Changes in Acceleration in Short Periods of Time, Night, Tyre Pressure C |

**Table A3.** CUPAC data subsets used in training WhONet

| CUPAC Dataset | Features | Total Time Driven, Distance Covered, Velocity and Acceleration |
|---|---|---|
| Alpha 1 | Inner-city, Parking lot, High traffic | 6.37 mins, 2.31 km, 0.518 to 39.35 km/hr, -0.25 to 0.16 g |
| Alpha 2 | Parking lot, Country road, Low | 8.33 mins, 4.46 km, 0.004 to 64.577 km/hr, -0.22 to 0.27 g |
| Alpha 2 | Inner-city, Country road, High traffic | 8.33 mins, 3.93 km, 0.0 to 55.174 km/hr, -0.21 to 0.21 g |
| Alpha 2 | Inner-city, High traffic | 8.33 mins, 3.99 km, 0.007 to 70.376 km/hr, -0.36 to 0.24 g |
| Bravo 1 | Inner-city, Medium traffic | 5.89 mins, 1.57 km, 0.0 to 40.374 km/hr, -0.27 to 0.16 g |
| Bravo 2 | Residential area, Road bumps, Low traffic | 16.92mins 7.46 km, 0.004 to 65.405 km/hr, -0.44 to 0.33 g |
| Bravo 3 | Residential area, Road bumps, Inner-city, Medium traffic | 25.38 mins, 12.51m, 0.0 to 69.48 km/hr, -0.68 to 0.31 g |
| Charlie 1 | Country road, Parking lot, Medium traffic | 4.86 mins,3.03 km, 0.004 to 62.68 km/hr, -0.26 to 0.21 g |
| Charlie 2 | Inner-city, Country road, Low traffic | 11.23 mins, 5.85 km, 0.004 to 58.104 km/hr, -0.38 to 0.27 g |
| Delta 1 | Highway, Residential area, Low traffic | 8.33 mins, 3.79 km, 0.004 to 53.654 km/hr, -0.19 to 0.25 g |

**Table A4.** IO-VNB data test subset used in the less challenging scenario. 1.75 min 2.64km

| Scenario | IO-VNB Data Subset | Total Time Driven, Distance covered, Velocity and Acceleration |
|---|---|---|
| Motorway | V-Vw12 | 1.75 min, 2.64 km, 82.6 to 97.4 km/hr, −0.06 to +0.07 g |

**Table A5.** IO-VNB data test subset used in the challenging scenarios.

| Challenging Scenarios | IO-VNB Data Subset | Total Time Driven, Distance Covered, Velocity and Acceleration |
|---|---|---|
| Roundabout | V-Vta11 | 1.0 min, 0.92 km, 26.8 to 97.7 km/hr, −0.45 to +0.15 g |
| | V-Vfb02d | 1.5 min, 0.84 km, 0.0 to 57.3 km/hr, −0.33 to +0.31 g |
| Changes in acceleration | V-Vfb02e | 1.6 min, 1.52 km, 37.4 to 73.9 km/hr, −0.24 to +0.19 g |
| | V-Vta12 | 1.0 mins, 1.27 km, 44.7 to 85.3 km/hr, 0.44 to +0.13 g |
| Hard Brake | V-Vw16b | 2.0 mins, 1.99 km, 1.3 to 86.3 km/hr, −0.75 to +0.29 g |
| | V-Vw17 | 0.5 min, 0.54 km, 31.5 to 72.7 km/hr, −0.8 to +0.19 g |
| | V-Vta9 | 0.4 min, 0.43 km, 48.9 to 87.7 km/hr, −0.6 to +0.14 g |
| Sharp Cornering and Successive left and right turns | V-Vw6 | 2.1 mins, 1.08 km, 3.3 to 40.7 km/hr, −0.34 to +0.26 g |
| | V-Vw7 | 2.8 mins, 1.23 km, 0.4 to 42.2 km/hr, −0.37 to +0.37 g |
| | V-Vw8 | 2.7 mins, 1.12 km, 0.0 to 46.4 km/hr, −0.37 to +0.27 g |
| Wet Road | V-Vtb8 | 1.2 mins, 1.35 km, 60.9 to 76.5 km/hr, -0.35 to 0.08 g |
| | V-Vtb11 | 0.7 min, 0.84 km, 65.1 to 75.3 km/hr, -0.05 to 0.12 g |
| | V-Vtb13 | 2.1 mins, 0.99 km, 7.5 to 43.3 km/hr, -0.31 to 0.22 g |

**Table A6.** IO-VNB data subsets used for the longer GNSS outage scenario performance evaluation of WhONet's

| IO-VNB Dataset | Features | Total Time Driven, Distance Covered, Velocity and Acceleration |
|---|---|---|
| V-Vtb3 | Reverse, Wet Road, Dirt Road, Gravel Road, Night-time, Tyre Pressure A | 13.8 mins, 0.71 km, 0.0 to 37.5 km/hr, -0.23 to 0.33 g |
| V_Vfb01c | Motorway (M62), wet-road, heavy traffic, nighttime, tyre pressure C | 10.5 mins, 10.66 km, 0.2 to 104.5 km/hr, -0.36 to 0.38 g |
| V_Vfb02a | Motorway (M1), roundabout (x2), A-road (A650), nighttime, hard-brakes, tyre pressure D | 59.9 mins, 96.5 km, 0.0 to 122.3 km/hr, -0.5 to 0.37 g |
| V_Vta1a | Wet road, gravel road, country road, sloppy roads, roundabout (x3), hard-brake on wet road, tyre pressure A | 43.0 mins, 40.74 km, 0.0 to 103.4 km/hr, -0.54 to 0.35 g |
| V_Vfb02b | Roundabout (x1), bumps, successive left-right turns, hard-brakes (x7), swift-maneuvers, nighttime, tyre pressure D | 18.3 mins, 7.69 km, 0.0 to 84.3 km/hr, -0.5 to 0.35 g |
| V_Vfb02g | Motorway (M1), A-road (A42, A444, A5), country road, roundabout (x2), hard-brakes, nighttime, tyre pressure D | 45.3 mins, 63.56 km, 0.0 to 119.4 km/hr, -0.51 to 0.35 g |
| V-St6 | Motorway(M40), daytime, tyre pressure E | 85.6 mins, 113.63 km, 0.0 to 122.1 km/hr, -0.32 to 0.35 g |
| V_St7 | Motorway (M40), residential roads, A-road (A46), tyre pressure E | 74.0 mins, 90.06 km, 0.0 to 117.9 km/hr, -0.3 to 0.3 g |
| V-S3a | Round-about (x15), u-turn/reverse drive (x4), motorway (M6), A-road (A4600, A426), hard-brake, swift maneuvers, country roads, change in speed, night-time, sharp turn left/right, tyre pressure E | 41.1 mins, 26.0 km, 0.0 to 98.0 km/hr, -0.57 to 0.4 g |

**Table A7.** CUPAC data subsets used for the longer GNSS outage scenario performance evaluation of WhONet's

| CUPAC Dataset | Total Time Driven, Distance Covered, Velocity and Acceleration |
|---|---|
| Charlie3 | 13.8 mins, 0.71 km, 0.004 to 77.252 km/hr, -0.42 to 0.25 g |
| Charlie 4 | 10.5 mins, 10.66 km, 0.007 to 58.694 km/hr, -0.34 to 0.31 g |
| Delta 2 | 59.9 mins, 96.5 km, 0.004 to 37.224 km/hr, -0.26 to 0.25 g |
| Delta 3 | 43.0 mins, 40.74 km, 0.004 to 60.75 km/hr, -0.36 to 0.25 g |

***Table A8.*** *Results from the 30 seconds GNSS outage experiment results on the IO-VNB dataset*

| IO-VNB Dataset | Performance Metrics | Position Estimation Error (m) | | | | Total Distance Travelled (m) | Number of Test Sequences evaluated |
|---|---|---|---|---|---|---|---|
| | | CTE | | CRSE | | | |
| | | Physics Model | WhONet | Physics Model | WhONet | | |
| V_Vtb3 | Max | 5.75 | 0.72 | 8.37 | 1.35 | 222 | 27 |
| | Min | 0.03 | 0.00 | 0.44 | 0.11 | 0 | |
| | μ | 0.84 | 0.23 | 2.31 | 0.67 | 24 | |
| | σ | 1.14 | 0.18 | 1.63 | 0.20 | 46 | |
| V_Vfb01c | Max | 9.62 | 1.38 | 17.73 | 2.41 | 830 | 21 |
| | Min | 0.20 | 0.00 | 2.19 | 0.22 | 50 | |
| | μ | 3.30 | 0.55 | 6.27 | 0.90 | 507 | |
| | σ | 2.46 | 0.40 | 3.36 | 0.43 | 259 | |
| V_Vfb02a | Max | 8.89 | 2.71 | 18.25 | 8.57 | 961 | 119 |
| | Min | 0.02 | 0.00 | 1.72 | 0.28 | 11 | |
| | μ | 1.04 | 0.43 | 3.73 | 1.01 | 807 | |
| | σ | 1.18 | 0.43 | 2.13 | 0.83 | 156 | |
| V_Vta1a | Max | 14.29 | 2.00 | 19.22 | 3.66 | 768 | 86 |
| | Min | 0.12 | 0.00 | 1.87 | 0.43 | 9 | |
| | μ | 3.66 | 0.42 | 8.24 | 1.15 | 473 | |
| | σ | 2.76 | 0.47 | 3.72 | 0.63 | 144 | |
| V_Vfb02b | Max | 11.60 | 0.88 | 17.04 | 1.70 | 570 | 36 |
| | Min | 0.12 | 0.00 | 4.46 | 0.37 | 47 | |
| | μ | 2.68 | 0.33 | 10.54 | 1.04 | 207 | |
| | σ | 2.73 | 0.23 | 2.94 | 0.30 | 109 | |
| V_Vfb02g | Max | 11.69 | 1.87 | 12.84 | 3.94 | 947 | 90 |
| | Min | 0.08 | 0.00 | 2.17 | 0.34 | 206 | |
| | μ | 1.78 | 0.38 | 4.97 | 0.94 | 706 | |
| | σ | 1.81 | 0.38 | 2.03 | 0.47 | 174 | |
| V_St6 | Max | 11.67 | 1.92 | 14.14 | 2.46 | 987 | 171 |
| | Min | 0.00 | 0.00 | 0.82 | 0.21 | 7 | |
| | μ | 2.40 | 0.52 | 3.89 | 0.93 | 664 | |
| | σ | 1.81 | 0.44 | 2.24 | 0.42 | 269 | |
| V_St7 | Max | 8.71 | 1.09 | 12.85 | 1.90 | 713 | 56 |
| | Min | 0.01 | 0.00 | 0.54 | 0.24 | 1 | |
| | μ | 2.56 | 0.31 | 5.28 | 0.80 | 337 | |
| | σ | 2.11 | 0.26 | 2.87 | 0.43 | 218 | |
| V_S3a | Max | 9.37 | 3.03 | 18.89 | 4.19 | 770 | 82 |
| | Min | 0.03 | 0.00 | 0.98 | 0.22 | 1 | |
| | μ | 3.14 | 0.55 | 7.44 | 1.22 | 316 | |
| | σ | 2.00 | 0.62 | 3.69 | 0.65 | 166 | |

***Table A9***. *Results from the 30 seconds GNSS outage experiment results on the CUPAC dataset*

| CUPAC Dataset | Performance Metrics | Position Estimation Error (m) | | | | Total Distance Travelled (m) | Number of Test Sequences evaluated |
|---|---|---|---|---|---|---|---|
| | | CTE | | CRSE | | | |
| | | Physics Model | WhONet | Physics Model | WhONet | | |
| Charlie 3 | Max | 10.50 | 2.34 | 14.52 | 5.36 | 629 | 45 |
| | Min | 0.02 | 0.00 | 2.44 | 0.37 | 113 | |
| | μ | 2.83 | 0.40 | 7.39 | 1.14 | 309 | |
| | *σ* | 2.56 | 0.48 | 2.75 | 0.73 | 151 | |
| Charlie 4 | Max | 7.68 | 5.16 | 17.47 | 8.53 | 441 | 22 |
| | Min | 0.41 | 0.00 | 2.77 | 0.41 | 30 | |
| | μ | 3.15 | 0.72 | 7.52 | 1.62 | 218 | |
| | *σ* | 1.93 | 1.02 | 3.11 | 1.69 | 97 | |
| Delta 2 | Max | 5.45 | 0.84 | 11.08 | 1.42 | 246 | 16 |
| | Min | 0.34 | 0.00 | 4.35 | 0.51 | 19 | |
| | μ | 2.24 | 0.34 | 7.35 | 0.94 | 188 | |
| | *σ* | 1.55 | 0.24 | 2.08 | 0.24 | 64 | |
| Delta 3 | Max | 6.81 | 1.29 | 13.26 | 5.07 | 358 | 38 |
| | Min | 0.00 | 0.00 | 2.54 | 0.43 | 6 | |
| | (μ | 1.88 | 0.36 | 7.13 | 1.18 | 130 | |
| | *σ* | 1.48 | 0.31 | 2.29 | 0.72 | 90 | |

***Table A10.*** *Results from the 60 seconds GNSS outage experiment results on the IO-VNB dataset*

| **IO-VNB Dataset** | **Performance Metrics** | **Position Estimation Error (m)** | | | | **Total Distance Travelled (m)** | **Number of Test Sequences evaluated** |
|---|---|---|---|---|---|---|---|
| | | **CTE** | | **CRSE** | | | |
| | | **Physics Model** | **WhONet** | **Physics Model** | **WhONet** | | |
| V_Vtb3 | Max | 2.53 | 0.62 | 15.04 | 2.29 | 358 | 13 |
| | Min | 0.03 | 0.00 | 0.95 | 0.29 | 1 | |
| | μ | 1.02 | 0.30 | 4.56 | 1.31 | 50 | |
| | *σ* | 0.68 | 0.16 | 3.22 | 0.25 | 90 | |
| V_Vfb01c | Max | 9.95 | 2.38 | 14.51 | 3.73 | 1557 | 10 |
| | Min | 0.15 | 0.00 | 5.97 | 0.58 | 149 | |
| | μ | 5.79 | 0.98 | 11.40 | 1.77 | 1042 | |
| | *σ* | 3.59 | 0.69 | 2.53 | 0.73 | 500 | |
| V_Vfb02a | Max | 8.68 | 2.43 | 31.34 | 9.62 | 1883 | 59 |
| | Min | 0.01 | 0.00 | 4.56 | 0.87 | 45 | |
| | μ | 1.42 | 0.66 | 7.48 | 2.02 | 1615 | |
| | *σ* | 1.63 | 0.56 | 3.81 | 1.11 | 310 | |
| V_Vta1a | Max | 17.45 | 3.02 | 30.94 | 6.27 | 1446 | 43 |
| | Min | 0.47 | 0.00 | 7.69 | 1.06 | 217 | |
| | μ | 6.23 | 0.67 | 16.47 | 2.29 | 947 | |
| | *σ* | 3.66 | 0.73 | 5.83 | 1.07 | 250 | |
| V_Vfb02b | Max | 7.73 | 0.96 | 30.67 | 2.94 | 1052 | 18 |
| | Min | 0.01 | 0.00 | 11.94 | 1.10 | 213 | |
| | μ | 2.38 | 0.39 | 21.08 | 2.08 | 415 | |
| | *σ* | 2.18 | 0.24 | 4.91 | 0.48 | 187 | |
| V_Vfb02g | Max | 7.94 | 2.16 | 22.41 | 4.42 | 1864 | 45 |
| | Min | 0.05 | 0.00 | 5.04 | 0.77 | 730 | |
| | μ | 2.49 | 0.62 | 9.95 | 1.88 | 1411 | |
| | *σ* | 2.22 | 0.54 | 3.72 | 0.65 | 334 | |
| V_St6 | Max | 11.88 | 3.30 | 26.03 | 4.32 | 1960 | 85 |
| | Min | 0.13 | 0.00 | 2.78 | 0.76 | 81 | |
| | µ | 3.88 | 0.90 | 7.77 | 1.86 | 1336 | |
| | $\sigma$ | 2.24 | 0.73 | 4.05 | 0.71 | 522 | |
| V_St7 | Max | 12.44 | 1.31 | 21.99 | 3.57 | 1393 | 28 |
| | Min | 0.15 | 0.00 | 1.57 | 0.59 | 2 | |
| | µ | 4.08 | 0.50 | 10.56 | 1.61 | 674 | |
| | $\sigma$ | 3.13 | 0.32 | 5.04 | 0.75 | 420 | |
| V_S3a | Max | 12.73 | 4.44 | 30.88 | 5.63 | 1514 | 41 |
| | Min | 0.08 | 0.00 | 3.37 | 0.83 | 8 | |
| | µ | 4.50 | 0.88 | 14.89 | 2.43 | 632 | |
| | $\sigma$ | 2.76 | 0.96 | 6.33 | 0.90 | 304 | |

***Table A11.*** *Results from the 60 seconds GNSS outage experiment results on the CUPAC dataset*

| **CUPAC Dataset** | **Performance Metrics** | **Position Estimation Error (m)** | | | | **Total Distance Travelled (m)** | **Number of Test Sequences evaluated** |
|---|---|---|---|---|---|---|---|
| | | **CTE** | | **CRSE** | | | |
| | | **Physics Model** | **WhONet** | **Physics Model** | **WhONet** | | |
| Charlie 3 | Max | 10.04 | 2.10 | 22.81 | 6.42 | 1238 | 22 |
| | Min | 0.29 | 0.00 | 5.88 | 1.06 | 280 | |
| | µ | 3.58 | 0.64 | 14.90 | 2.28 | 622 | |
| | $\sigma$ | 2.46 | 0.57 | 4.74 | 1.01 | 295 | |
| Charlie 4 | Max | 8.52 | 3.87 | 23.03 | 9.59 | 669 | 11 |
| | Min | 0.13 | 0.00 | 10.28 | 1.01 | 169 | |
| | µ | 3.88 | 1.17 | 15.04 | 3.23 | 435 | |
| | $\sigma$ | 2.71 | 1.02 | 3.42 | 2.64 | 145 | |
| Delta 2 | Max | 6.24 | 0.96 | 19.22 | 2.48 | 481 | 8 |
| | Min | 0.37 | 0.00 | 10.51 | 1.25 | 153 | |
| | µ | 2.53 | 0.56 | 14.70 | 1.88 | 376 | |
| | $\sigma$ | 1.63 | 0.21 | 3.07 | 0.39 | 100 | |
| Delta 3 | Max | 6.13 | 1.53 | 20.40 | 5.99 | 625 | 16 |
| | Min | 0.12 | 0.00 | 8.75 | 1.51 | 23 | |
| | µ | 2.56 | 0.49 | 14.25 | 2.36 | 261 | |
| | $\sigma$ | 1.77 | 0.45 | 2.84 | 0.97 | 165 | |

***Table A12.*** *Results from the 120 seconds GNSS outage experiment results on the IO-VNB dataset*

| **IO-VNB Dataset** | **Performance Metrics** | **Position Estimation Error (m)** | | | | **Total Distance Travelled (m)** | **Number of Test Sequences evaluated** |
|---|---|---|---|---|---|---|---|
| | | **CTE** | | **CRSE** | | | |
| | | **Physics Model** | **WhONet** | **Physics Model** | **WhONet** | | |
| V_Vtb3 | Max | 2.67 | 0.63 | 15.99 | 4.01 | 359 | 6 |
| | Min | 0.61 | 0.01 | 6.74 | 1.75 | 25 | |
| | µ | 1.78 | 0.30 | 9.11 | 2.62 | 102 | |
| | $\sigma$ | 0.66 | 0.21 | 3.21 | 0.19 | 117 | |
| V_Vfb01c | Max | 19.55 | 3.18 | 27.18 | 5.05 | 3058 | 5 |
| | Min | 3.54 | 0.01 | 19.23 | 1.95 | 543 | |
| | µ | 11.26 | 1.41 | 22.80 | 3.54 | 2083 | |
| | $\sigma$ | 5.13 | 0.90 | 3.42 | 0.67 | 888 | |
| V_Vfb02a | Max | 9.79 | 2.24 | 46.50 | 11.75 | 3671 | 29 |
| | Min | 0.10 | 0.00 | 9.91 | 2.05 | 683 | |
| | µ | 1.90 | 0.89 | 15.04 | 4.07 | 3234 | |
| | σ | 2.13 | 0.63 | 6.66 | 1.41 | 577 | |
| V_Vta1a | Max | 19.61 | 1.99 | 51.34 | 10.35 | 2579 | 21 |
| | Min | 1.26 | 0.00 | 20.26 | 2.64 | 785 | |
| | μ | 11.46 | 0.79 | 33.06 | 4.62 | 1893 | |
| | σ | 4.59 | 0.56 | 9.53 | 1.81 | 430 | |
| V_Vfb02b | Max | 7.42 | 1.26 | 56.24 | 5.76 | 1489 | 9 |
| | Min | 0.44 | 0.00 | 27.49 | 2.52 | 514 | |
| | μ | 2.59 | 0.68 | 42.17 | 4.15 | 830 | |
| | σ | 2.33 | 0.34 | 8.66 | 0.88 | 288 | |
| V_Vfb02g | Max | 8.55 | 2.53 | 34.39 | 5.66 | 3716 | 22 |
| | Min | 0.00 | 0.00 | 11.34 | 2.13 | 1707 | |
| | μ | 3.79 | 1.11 | 19.48 | 3.79 | 2852 | |
| | σ | 2.64 | 0.73 | 5.66 | 0.83 | 581 | |
| V_St6 | Max | 14.95 | 5.13 | 35.61 | 6.39 | 3870 | 42 |
| | Min | 0.75 | 0.00 | 7.49 | 1.63 | 256 | |
| | μ | 6.97 | 1.52 | 15.44 | 3.69 | 2700 | |
| | σ | 3.52 | 1.31 | 7.01 | 1.07 | 977 | |
| V_St7 | Max | 18.22 | 1.73 | 36.17 | 6.49 | 2435 | 14 |
| | Min | 0.57 | 0.00 | 11.75 | 1.43 | 216 | |
| | μ | 7.35 | 0.68 | 21.13 | 3.21 | 1348 | |
| | σ | 4.69 | 0.45 | 7.34 | 1.23 | 784 | |
| V_S3a | Max | 15.22 | 4.81 | 51.50 | 8.07 | 2623 | 20 |
| | Min | 0.94 | 0.00 | 13.18 | 2.37 | 422 | |
| | μ | 7.84 | 1.43 | 29.89 | 4.82 | 1269 | |
| | σ | 3.97 | 1.31 | 10.24 | 1.26 | 558 | |

***Table A13.*** *Results from the 120 seconds GNSS outage experiment results on the CUPAC dataset*

| **CUPAC Dataset** | **Performance Metrics** | **Position Estimation Error (m)** | | | | **Total Distance Travelled (m)** | **Number of Test Sequences evaluated** |
|---|---|---|---|---|---|---|---|
| | | **CTE** | | **CRSE** | | | |
| | | **Physics Model** | **WhONet** | **Physics Model** | **WhONet** | | |
| Charlie 3 | Max | 17.82 | 2.27 | 36.18 | 9.40 | 2339 | 11 |
| | Min | 2.05 | 0.00 | 14.78 | 2.27 | 681 | |
| | μ | 6.95 | 0.92 | 29.80 | 4.57 | 1245 | |
| | σ | 4.34 | 0.58 | 5.94 | 1.61 | 510 | |
| Charlie 4 | Max | 9.18 | 2.91 | 37.42 | 12.10 | 1229 | 5 |
| | Min | 3.50 | 0.00 | 25.21 | 2.60 | 612 | |
| | μ | 7.56 | 1.83 | 31.02 | 6.65 | 923 | |
| | σ | 2.07 | 0.52 | 4.13 | 3.71 | 214 | |
| Delta 2 | Max | 8.54 | 0.85 | 37.15 | 4.68 | 850 | 4 |
| | Min | 1.95 | 0.00 | 26.23 | 2.70 | 634 | |
| | μ | 5.06 | 0.46 | 29.41 | 3.75 | 752 | |
| | σ | 2.47 | 0.06 | 4.49 | 0.45 | 78 | |
| Delta 3 | Max | 7.93 | 2.10 | 37.73 | 5.74 | 990 | 9 |
| | Min | 0.29 | 0.00 | 22.02 | 3.15 | 173 | |
| | μ | 3.73 | 0.77 | 28.49 | 4.30 | 481 | |
| | σ | 2.27 | 0.49 | 4.56 | 0.66 | 263 | |

***Table A14.*** *Results from the 180 seconds GNSS outage experiment on the IO-VNB dataset*

| **IO-VNB Dataset** | **Position Estimation Error (m)** | | **Total Distance** | **Number of Test** |
|---|---|---|---|---|
| | **CTE** | **CRSE** | | |

| | Performance Metrics | Physics Model | WhONet | Physics Model | WhONet | Travelled (m) | Sequences evaluated |
|---|---|---|---|---|---|---|---|
| V_Vtb3 | Max | 4.26 | 1.02 | 18.29 | 4.97 | 369 | 4 |
| | Min | 0.85 | 0.01 | 10.32 | 2.81 | 46 | |
| | μ | 2.30 | 0.49 | 13.67 | 3.93 | 153 | |
| | σ | 1.25 | 0.18 | 2.89 | 0.21 | 127 | |
| V_Vfb01c | Max | 21.97 | 0.38 | 36.12 | 7.42 | 4407 | 3 |
| | Min | 3.39 | 0.00 | 31.92 | 3.97 | 1835 | |
| | μ | 15.53 | 0.26 | 33.57 | 5.43 | 3340 | |
| | σ | 8.59 | 0.01 | 1.82 | 0.66 | 1095 | |
| V_Vfb02a | Max | 12.63 | 3.02 | 52.18 | 13.02 | 5447 | 19 |
| | Min | 0.03 | 0.00 | 15.67 | 3.67 | 1102 | |
| | μ | 2.28 | 1.17 | 22.72 | 6.08 | 4858 | |
| | σ | 3.02 | 0.85 | 8.21 | 1.55 | 915 | |
| V_Vta1a | Max | 27.18 | 1.68 | 73.89 | 14.19 | 3692 | 14 |
| | Min | 10.71 | 0.00 | 31.29 | 4.20 | 1173 | |
| | μ | 17.19 | 0.79 | 49.59 | 6.93 | 2840 | |
| | σ | 4.41 | 0.48 | 10.84 | 2.42 | 661 | |
| V_Vfb02b | Max | 5.63 | 1.10 | 75.46 | 8.09 | 1910 | 6 |
| | Min | 0.90 | 0.00 | 55.23 | 4.04 | 890 | |
| | μ | 3.17 | 0.56 | 63.25 | 6.23 | 1244 | |
| | σ | 1.81 | 0.38 | 8.01 | 0.77 | 361 | |
| V_Vfb02g | Max | 9.28 | 2.53 | 53.40 | 8.33 | 5251 | 15 |
| | Min | 0.09 | 0.00 | 18.65 | 3.48 | 2474 | |
| | μ | 3.55 | 1.27 | 29.84 | 5.65 | 4234 | |
| | σ | 2.57 | 0.76 | 9.48 | 0.89 | 916 | |
| V_St6 | Max | 21.86 | 6.67 | 41.30 | 8.62 | 5626 | 28 |
| | Min | 0.62 | 0.00 | 11.42 | 2.61 | 513 | |
| | μ | 10.37 | 2.14 | 23.16 | 5.53 | 4050 | |
| | σ | 5.25 | 1.64 | 9.46 | 1.24 | 1412 | |
| V_St7 | Max | 19.09 | 1.67 | 54.62 | 9.47 | 3570 | 9 |
| | Min | 0.23 | 0.00 | 19.00 | 2.54 | 610 | |
| | μ | 11.10 | 0.84 | 31.82 | 4.89 | 2000 | |
| | σ | 6.31 | 0.40 | 11.06 | 1.62 | 1083 | |
| V_S3a | Max | 26.28 | 4.62 | 64.02 | 11.14 | 3345 | 13 |
| | Min | 0.20 | 0.00 | 20.07 | 4.32 | 1086 | |
| | μ | 11.59 | 1.65 | 44.75 | 7.23 | 1906 | |
| | σ | 6.34 | 1.08 | 12.72 | 1.37 | 775 | |

***Table A25.*** *Results from the 180 seconds GNSS outage experiment on the CUPAC dataset*

| CUPAC Dataset | Performance Metrics | Position Estimation Error (m) | | | | Total Distance Travelled (m) | Number of Test Sequences evaluated |
|---|---|---|---|---|---|---|---|
| | | CTE | | CRSE | | | |
| | | Physics Model | WhONet | Physics Model | WhONet | | |
| Charlie 3 | Max | 17.10 | 1.50 | 52.56 | 11.04 | 3416 | 7 |
| | Min | 3.55 | 0.00 | 23.04 | 3.89 | 1134 | |
| | μ | 10.33 | 1.03 | 44.08 | 6.91 | 1899 | |
| | σ | 4.54 | 0.28 | 10.46 | 1.77 | 788 | |
| Charlie 4 | Max | 15.11 | 4.34 | 55.23 | 20.79 | 1603 | 3 |
| | Min | 8.03 | 0.00 | 40.66 | 3.70 | 1171 | |
| | μ | 11.72 | 2.89 | 46.80 | 10.28 | 1422 | |
| | σ | 2.90 | 0.20 | 6.17 | 7.00 | 183 | |
| Delta 2 | Max | 10.93 | 0.57 | 53.53 | 5.15 | 1214 | 2 |
| | Min | 5.56 | 0.00 | 37.87 | 5.06 | 1162 | |
| | μ | 8.24 | 0.51 | 45.70 | 5.13 | 1188 | |
| | $\sigma$ | 2.69 | 0.01 | 7.83 | 0.00 | 26 | |
| Delta 3 | Max | 10.17 | 2.10 | 50.38 | 7.55 | 1244 | 6 |
| | Min | 0.41 | 0.00 | 30.77 | 5.42 | 301 | |
| | μ | 4.59 | 1.18 | 42.74 | 6.45 | 721 | |
| | $\sigma$ | 3.42 | 0.16 | 7.22 | 0.54 | 346 | |